\documentclass[fleqn,10pt]{wlscirep}
\usepackage[utf8]{inputenc}
\usepackage[T1]{fontenc}
\usepackage{subcaption} 
\usepackage{caption}
\usepackage{graphicx}
\usepackage{longtable}
\usepackage{rotating}
\usepackage{booktabs}
\usepackage{multirow}

\usepackage{hyperref}
\usepackage{xr-hyper}

\usepackage{tikz}
\usetikzlibrary{positioning, arrows.meta, fit, backgrounds}

\title{Markerless Motion Capture in Routine Clinical Upper Limb Assessments: Validity and Insights Beyond Ordinal Scoring}

\author[1,2,*]{Tim Unger}
\author[2]{Olivier Lambercy}
\author[2]{Roger Gassert}
\author[3,4]{Andreas R. Luft}
\author[5]{R. James Cotton}
\author[1]{Chris Easthope Awai}

\affil[1]{Data Analytics \& Rehabilitation Technology (DART), Lake Lucerne Institute, Vitznau, Switzerland}
\affil[2]{Rehabilitation Engineering Laboratory, ETH Zurich, Zurich, Switzerland}
\affil[3]{Lake Lucerne Institute, Vitznau, Switzerland}
\affil[4]{cereneo, Vitznau, Switzerland}
\affil[5]{Shirley Ryan AbilityLab, Department of Physical Medicine and Rehabilitation, Northwestern University, Chicago, IL, USA}

\affil[*]{tim.unger@llui.org}

\begin{abstract}
The Action Research Arm Test (ARAT) is a widely-used upper limb outcome 
measure in neurorehabilitation, but its ordinal scoring is subjective and 
suffers from limited sensitivity and specificity. We evaluated whether 
artificial-intelligence (AI)-based markerless motion capture (MMC), embedded into ARAT assessments 
during clinical routine, accurately reconstructs upper limb 
movement and yields valid, objective kinematic metrics carrying clinically 
meaningful information beyond the ordinal score. Across 47 sessions from 20 
mixed-neurological patients (1,174 ARAT tasks), biomechanical reconstruction 
was accurate and robust across impairment levels, and kinematic metrics 
showed the discrimination pattern expected of a construct-valid measure. In 
longitudinal case studies, the metrics added the specificity and sensitivity 
the ordinal score lacks: a domain decomposition exposed patient-specific 
recovery profiles underlying equal ARAT gains (specificity), and kinematic 
improvement continued to be detected after the ARAT had saturated 
(sensitivity). MMC in clinical routine can thus provide valid, objective, 
sensitive, and specific kinematic measurement complementing ordinal scoring.
\end{abstract}

\begin{document}

\flushbottom
\maketitle
%
%
\thispagestyle{empty}


\section*{Introduction}

Neurological conditions such as stroke and Parkinson's disease affect 
hundreds of millions of people worldwide, making them a leading cause of 
motor disability \cite{gbd_2021_nervous_system_disorders_collaborators_global_2024}. 
Every year, more than 12 million people suffer a stroke \cite{feigin_world_2025}, 
and approximately half of the survivors still experience upper limb impairment after 
rehabilitation \cite{nakayama_recovery_1994}, often limiting their 
independence in the long term 
\cite{gbd_2021_nervous_system_disorders_collaborators_global_2024}. 
With aging societies, the number of affected individuals is expected 
to rise, increasing pressure on healthcare systems 
\cite{feigin_world_2025} and demanding more effective and efficient 
rehabilitation strategies \cite{kwakkel_standardized_2019, 
kwakkel_standardized_2017}. Precision neurorehabilitation has been 
proposed as the framework to meet this demand: a closed loop in which 
interventions are continually adapted to a patient's measured response 
and outcomes re-assessed to steer the next decision. Realising it 
depends on the quality of the measurements that close the loop, yet 
the outcome measures available in routine practice remain largely 
subjective, human-scored assessments that characterise functional 
state only coarsely. Outcome measures that capture functional state objectively, 
sensitively, and specifically are therefore a prerequisite for 
precision neurorehabilitation and for the development of improved 
surrogate biomarkers of recovery against which novel interventions can 
be evaluated.

When it comes to motor recovery, it is currently not feasible to measure 
movement objectively in clinical routine, and hence outcome measures rely on 
therapists observing the patient performing standardised tasks and then 
scoring them on an ordinal scale based on their visual perception. The 
Action Research Arm Test (ARAT) is one of the most frequently applied 
measures in neurorehabilitation to assess upper limb function in patients 
\cite{van_der_lee_action_2024}. It is a 19-item measure assessing upper limb 
functional performance on pick-and-place tasks (grasp, grip, pinch) and 
gross motor tasks, each rated on a four-point ordinal scale: 0 indicating 
inability to perform any part of the task, 1 indicating partial performance, 
2 indicating task completion but with increased time or difficulty, and 3 
indicating normal motor-performance \cite{van_der_lee_action_2024}. The ARAT's 
ordinal scale introduces several inter-related limitations 
\cite{demers_activity_2017, kwakkel_standardized_2019, pike_systematic_2018}. 
First, subjectivity \cite{van_der_lee_action_2024, 
nordin_intra-rater_2014}: distinguishing between Score~2 and Score~3 
relies on a clinician's judgement of whether movement is ``normal'', a 
determination that is inherently subjective and susceptible to 
inter-rater variability, particularly at the functional boundary where 
movement quality is marginal. Despite continued efforts to harmonise 
scoring criteria across neurorehabilitation clinics \cite{prange-lasonder_european_2021, yozbatiran_standardized_2008, burridge_systematic_2019, herve-colas_standardized_2026}, residual 
between-rater and between-site differences persist, and reliable 
scoring itself requires training and accrues over a learning curve. Second, sensitivity 
\cite{wilson_analysing_2021}: like many ordinal clinical scales, the ARAT 
suffers from floor and ceiling effects, meaning patients at the extremes of 
performance cannot be further differentiated by the scale, and meaningful 
improvements in movement quality or speed may occur without any change in 
ordinal score. Third, specificity \cite{demers_activity_2017}: an ordinal 
score conveys no information about why a particular rating was assigned. Two 
patients receiving Score~2 on a specific task may do so for entirely different reasons (one 
constrained by limited range of motion, another by slow and effortful 
movement), yet the ARAT provides no means to distinguish these distinct 
movement quality profiles, limiting its value for treatment planning and 
outcome monitoring.

Kinematic analysis of upper limb movement offers a principled route to 
address these limitations, providing continuous, observer-independent 
measurements of movement quality. Marker-based optical motion capture (OMC) 
has demonstrated this potential clearly: kinematic metrics derived from an 
instrumented drinking task can quantify upper limb movement quality in 
people with stroke with high sensitivity and validity 
\cite{alt_murphy_kinematic_2018, murphy_responsiveness_2013}. However, 
OMC's substantial time requirements, technical complexity, and cost render 
it incompatible with clinical routine, limiting its role to occasional 
laboratory-based assessments rather than integrated rehabilitation 
monitoring. A practical kinematic measurement solution, one that is both 
accurate and minimally disruptive to clinical workflows, is therefore needed 
to bridge the gap.

Markerless motion capture (MMC), an emerging technology, derives 
full-body kinematics from multi-camera red--green--blue (RGB) video 
using AI-based computer vision, supported by a growing ecosystem of 
research and commercial tools 
\cite{cotton_differentiable_2025, pagnon_pose2sim_2022, dsouza_comparison_2024, matthis_freemocap_2026, uhlrich_opencap_2023, cuevas-velasquez_mamma_2026}. 
Rather than tracking physical markers attached to the body, as in 
active or passive marker-based systems, it detects anatomical 
landmarks directly in the image using pose estimation, removing the 
need for any on-body instrumentation. This 
approach has shown high accuracy in laboratory validation against 
marker-based optical motion capture \cite{unger_differentiable_2025} and 
minimal disruption to clinical workflows \cite{unger_towards_2026}. While 
these prior contributions establish that MMC can be accurately reconstructed 
under controlled laboratory conditions and integrated into clinical workflow 
without disruption, the present study takes the next step: evaluating 
whether the kinematic metrics collected during routine clinical sessions are 
themselves valid and clinically informative.

Evaluating MMC-based kinematics in this context calls for an incremental 
strategy, building from the underlying reconstruction up to the metrics 
derived from it and, ultimately, to the clinical value those metrics offer. 
The starting point is the kinematic reconstruction itself: before any metric 
can be trusted, the reconstructed joint trajectories must be of adequate 
accuracy in the clinical environment, and this accuracy must hold across the 
full range of patient impairment rather than degrading where motor function 
is worst.

Once an accurate signal is established, the central question becomes whether 
metrics derived from it validly capture the upper limb function construct. 
This is methodologically non-trivial because no gold-standard continuous 
measure of upper limb function exists in clinical routine: validation must 
proceed against the very ordinal system whose limitations motivate the work. 
This apparent circularity can be turned into an advantage by asking not 
simply whether kinematic metrics agree with the ARAT, but whether they agree 
\emph{in the way a valid measure should}. A metric that genuinely captures 
upper limb function should discriminate strongly between score groups where 
the clinical boundary is robust, yet discriminate less where the ordinal 
scale is itself unreliable. Observing this differential pattern would 
provide evidence of construct validity that simple agreement cannot, 
precisely because it tracks the underlying functional construct rather than 
the ordinal, clinical scale.

A metric established as valid in this cross-sectional sense can then be 
examined longitudinally, both to confirm that its validity holds as patients 
recover over time and to expose the added clinical value it offers beyond 
the ordinal score: the ability to track recovery continuously, to reveal 
which specific aspects of movement drive a given change, and to remain 
sensitive to improvement even after the clinical scale has saturated. 
Together, these strands of evidence would establish whether MMC-derived 
kinematics can serve as a valid and clinically informative complement to 
ordinal scoring.

The aim of the present study is to evaluate the validity and added clinical 
value of AI-based markerless motion capture for upper limb assessment in 
routine neurorehabilitation, by quantifying the accuracy of the underlying 
kinematic reconstruction, the construct validity of metrics derived from it 
at the population and longitudinal levels, and the clinical information 
those metrics carry beyond ordinal scoring.

We therefore test the following hypotheses (Table~\ref{tab:framework}):

\textbf{H1 (Reconstruction accuracy and robustness).} Biomechanical 
reconstruction from an ad-hoc three-camera setup (see Supplementary Figure \ref{fig:ARAT_camera_setup}) yields 
kinematic time series of adequate accuracy for metric extraction 
(cohort-level mean reprojection error within the 10--20\,pixel (px) range 
previously established as acceptable for this reconstruction algorithm 
in clinical use, modestly above the 5--10\,px laboratory standard 
\cite{unger_differentiable_2025}), with reconstruction quality 
remaining stable across the full range of clinical impairment (effect 
of ARAT score group on reprojection error $\eta^2_H < 0.06$, 
conventional small-effect threshold).

\textbf{H2 (Construct validity at the population level).} MMC-derived 
kinematic metrics capture the intended movement-quality construct, 
evidenced by Tier~1 (Score~0/1 vs.\ Score~2/3) combined-feature 
discrimination of area under the receiver operating characteristic 
curve (AUC) $\geq 0.85$ (conventional ``excellent 
discrimination'' threshold \cite{hosmer_applied_2013}) in the 
majority of analysis groups, with Tier~2 (Score~2 vs.\ Score~3) 
discrimination remaining above chance (AUC $> 0.5$, exceeding the 
conventional ``acceptable discrimination'' threshold of 0.7 
\cite{hosmer_applied_2013} where possible) but consistently lower 
than Tier~1 across analysis groups, the pattern expected of a valid 
construct measure that tracks the underlying function rather than 
the noisy scale.

\textbf{H3a (Longitudinal construct validity).} During sub-ceiling 
recovery, the overall kinematic compound score $S_{\mathrm{overall}}$, 
which aggregates all kinematic metrics, tracks the clinical ARAT 
trajectory within the adopted minimal clinically important difference 
(mean absolute deviation, MAD $<$ 15\,percentage points (pp) on the 
normalised compound scale, one minimal clinically important difference (MCID) 
\cite{kwakkel_standardized_2019}), evaluated per analysis group, 
establishing longitudinal construct validity in each analysis group where 
the criterion holds. Calculation of the metrics is described in 
Figure~\ref{fig:compound_score_aggregation}.

\textbf{H3b (Specificity: domain-level recovery profiles beyond the 
ordinal score).} Given the longitudinal validity established in H3a, the 
domain decomposition into range-of-motion (ROM), velocity, and compensation 
scores ($S_{\mathrm{ROM}}$, $S_{\mathrm{Vel}}$, $S_{\mathrm{Comp}}$) 
exposes patient-specific recovery profiles invisible to the ordinal 
score when they exist. The case studies illustrate this added clinical 
value through the observed contrast between two patients on the Grasp 
analysis group.

\textbf{H3c (Responsiveness beyond the clinical ceiling).} Given the 
validity established in H3a, after clinical scores plateau at the ARAT 
ceiling, at least one domain-level compound score continues to change 
by more than the adopted MCID 
($\geq$ 15\,pp on the normalised compound scale, one MCID 
\cite{kwakkel_standardized_2019}) during the post-ceiling window, 
demonstrating responsiveness where ordinal scoring saturates.

H1 and H2 are tested at the cohort level and constitute the primary 
hypotheses. H3a--H3c are evaluated through two longitudinal case 
studies and are exploratory, intended to illustrate the added clinical 
value of kinematic metrics and to motivate future cohort-level 
investigation. H3b and H3c each build on H3a: the longitudinal 
validity established in H3a is what licenses the domain divergence 
(H3b) and the post-ceiling change (H3c) to be read as genuine recovery 
rather than measurement noise; H3b and H3c are otherwise independent 
properties of the validated measure. 

Demonstrating these properties together would 
establish AI-based markerless motion capture as a valid, clinically 
meaningful complement to ordinal scoring inside an unmodified 
neurorehabilitation routine, opening a path toward continuous, 
multi-dimensional, observer-independent characterisation of upper 
limb recovery in standard clinical practice.

\begin{table}[ht]
\caption{Evaluation framework. Each hypothesis provides a distinct 
level of evidence, building from the accuracy of the underlying 
kinematic reconstruction from the AI-based markerless motion capture 
algorithm (H1), through construct validity of kinematic metrics at 
the population (H2) and longitudinal (H3a) levels, to the added 
clinical value of kinematic metrics beyond ordinal scoring (H3b, H3c). 
Threshold symbols: $\eta^2_H$, Kruskal--Wallis effect size 
(small-effect cut-off $0.06$); AUC, area under the ROC curve 
(``excellent'' $\geq 0.85$, ``acceptable'' $\geq 0.7$ 
\cite{hosmer_applied_2013}); MCID, minimal clinically important 
difference ($15$\,pp on the compound scale), adopted as an approximate 
threshold translated from consensus recommendations 
\cite{kwakkel_standardized_2019}. Operational definitions are given in 
Methods.}
\label{tab:framework}
\centering
\small
\renewcommand{\arraystretch}{1.3}
\begin{tabular}{p{0.5cm} p{2.6cm} p{3.6cm} p{2.3cm} p{2.5cm} p{2.4cm}}
\hline\hline
 & \textbf{Claim} & \textbf{Evidence type / method} & \textbf{Level} & \textbf{Data} & \textbf{Threshold} \\
\hline
\textbf{H1}   & Reconstruction accuracy \& robustness & Reprojection error vs.\ established range; stability across score groups & Trial (precondition) & Cohort (1,174 trials) & $\overline{\text{error}} \in [10,20]$\,px AND $\eta^2_H < 0.06$ \\
\textbf{H2}   & Construct validity (population) & Known-groups discrimination (Tier~1 strong, Tier~2 attenuated) & Population (cross-sectional) & Cohort ($n=20$) & Tier~1 AUC $\geq 0.85$ (majority); $0.5 <$ Tier~2 $<$ Tier~1 \\
\textbf{H3a} & Longitudinal construct validity & $S_{\mathrm{overall}}$ agrees with clinical change (MAD) & Longitudinal (within-patient) & Case studies ($n=2$) & $\mathrm{MAD}(S_{\mathrm{overall}}) <$ MCID, per analysis group \\
\textbf{H3b} & Specificity (added value) & Domain-level recovery profiles where present & Longitudinal & Case studies ($n=2$) & Reported, not a pass criterion \\
\textbf{H3c}  & Responsiveness beyond ceiling (added value) & Domain change after clinical ceiling reached & Longitudinal & Case study (n=1) & $\Delta S_{\mathrm{domain}} > $ MCID for $\geq 1$ domain (post-ceiling)\\
\hline\hline
\end{tabular}
\end{table}

\newpage
\section*{Methods}
\label{sec:methods}

\subsection*{Participants and Data Collection}
\label{sec:participants}

Data were collected from 20 participants (10 stroke, 6 Parkinson's 
disease, 4 other neurological conditions) during routinely scheduled 
ARAT sessions at the Swiss neurorehabilitation clinic cereneo, yielding 47 
sessions and 1,304 raw ARAT tasks (trials) (1,174 retained after outlier 
exclusion) in total. Three webcams (Logitech Brio 4K: 
1280 $\times$ 720 px, 60\,Hz) were set up ad-hoc around the assessment 
table (see Supplementary Figure \ref{fig:ARAT_camera_setup}), while therapists used a tablet application to mark start and 
stop of each individual trial and assign clinical ARAT scores in 
real time. All participants provided informed consent for research 
use of their data; ethics approval was granted by the local ethics 
committee (EKNZ, 2024-00196).

\subsection*{Kinematic Reconstruction and Preprocessing}
\label{sec:processing}

Upper limb kinematics were reconstructed using a differentiable 
biomechanics algorithm \cite{cotton_differentiable_2025}, previously 
validated against gold-standard OMC for upper limb reconstruction 
in people with stroke \cite{unger_differentiable_2025}. The 
resulting joint angle, angular velocity, wrist and trunk position 
time series were low-pass filtered using a zero-phase fourth-order 
Butterworth filter at 5\,Hz to remove high-frequency noise while 
preserving physiologically relevant movement dynamics. Trials were 
excluded if any derived metric fell outside three times the 
interquartile range of its respective ARAT score group 
(IQR\,$\times$\,3.0).

\subsection*{Data Quality of Kinematic Reconstruction}

Reconstruction quality was quantified using the mean reprojection 
error, which is commonly used as an internal measure of 
reconstruction quality in markerless motion capture systems 
\cite{cotton_differentiable_2025, unger_differentiable_2025}. It is 
defined as the Euclidean distance in pixels (px) between the detected 2D 
keypoints in each camera view and the reprojection of the 
corresponding reconstructed 3D joint positions back onto that view. 
For each trial, the reprojection error was averaged across all 
keypoints, camera views, and frames to yield a single trial-level 
quality measure, with lower values indicating closer agreement 
between the biomechanical reconstruction and the observed image 
data.

To evaluate H1, we defined two criteria for adequate and robust 
reconstruction. First, for \emph{adequacy}, the cohort-level mean 
reprojection error had to fall within the 10--20\,px range 
previously established as acceptable for this reconstruction 
algorithm in clinical use, modestly above the 5--10\,px laboratory 
baseline \cite{unger_differentiable_2025}; values within this range 
indicate reconstruction quality comparable to controlled 
clinical-routine conditions, while substantially larger errors 
would indicate inadequate quality. Second, for \emph{robustness 
across impairment}, reconstruction quality had to be stable across 
the range of clinical ability, such that reprojection error did not 
depend systematically on ARAT score; a large, systematic increase 
in error with greater impairment would indicate 
performance-dependent reconstruction bias and reject this criterion. 
Reprojection error was compared across the four ARAT score groups 
using a Kruskal--Wallis test, with $\eta^2_H$ as the effect size 
and a negligible-to-small effect ($\eta^2_H < 0.06$) taken as 
consistent with robustness; pairwise Dunn's post-hoc tests with 
Holm correction and rank-biserial correlation $r$ characterised 
any group differences. H1 was considered supported if both the 
adequacy and robustness criteria were met.

\subsection*{Kinematic Metrics}

\begin{table}[ht]
\caption{Standard ARAT task grouping and analysis groups used in the present study.
Tasks sharing the same gross motor movement pattern were pooled for analysis;
tasks with a unique movement pattern were analysed individually.}
\label{tab:task_grouping}
\centering
\begin{tabular}{ l l l }
\hline\hline
ARAT Subtest & Task & Analysis Group \\
\hline
\multirow{6}{*}{Grasp}
  & Block 10\,cm          & \multirow{6}{*}{Grasp (pooled)} \\
  & Block 7.5\,cm         & \\
  & Block 5\,cm           & \\
  & Block 2.5\,cm         & \\
  & Stone                 & \\
  & Ball 7.5\,cm          & \\
\hline
\multirow{4}{*}{Grip}
  & Water pouring         & Water Pouring (individual) \\
  \cline{2-3}
  & Tube 2.25\,cm         & \multirow{3}{*}{Grip (pooled)} \\
  & Tube 1\,cm            & \\
  & Washer                & \\
\hline
\multirow{6}{*}{Pinch}
  & Ball 6\,mm (index)    & \multirow{6}{*}{Pinch (pooled)} \\
  & Ball 6\,mm (middle)   & \\
  & Ball 6\,mm (ring)     & \\
  & Ball 1.5\,cm (index)  & \\
  & Ball 1.5\,cm (middle) & \\
  & Ball 1.5\,cm (ring)   & \\
\hline
\multirow{3}{*}{\shortstack[l]{Gross\\Movement}}
  & Hand behind head      & Hand Behind Head (individual) \\
  & Hand to mouth         & Hand to Mouth (individual) \\
  & Hand on top of head   & Hand on Top of Head (individual) \\
\hline\hline
\end{tabular}
\end{table}

\subsubsection*{Metric Selection}

Movement-quality metric sets with established psychometric evidence 
in neurological populations currently exist for the drinking task 
\cite{alt_murphy_kinematic_2018, murphy_responsiveness_2013}, 
but no equivalent evidence-based metric set exists for the ARAT. 
We therefore drew on the broader upper-limb kinematic literature, 
which has converged on a shared time-series substrate of joint 
angles, peak joint and end-effector (hand) velocities, and trunk 
displacement as a compensation indicator 
\cite{schwarz_systematic_2019, kwakkel_standardized_2019}, to 
construct a general-purpose metric set designed to represent 
overall movement quality across all seven ARAT analysis groups 
rather than to be optimised for any one of them.

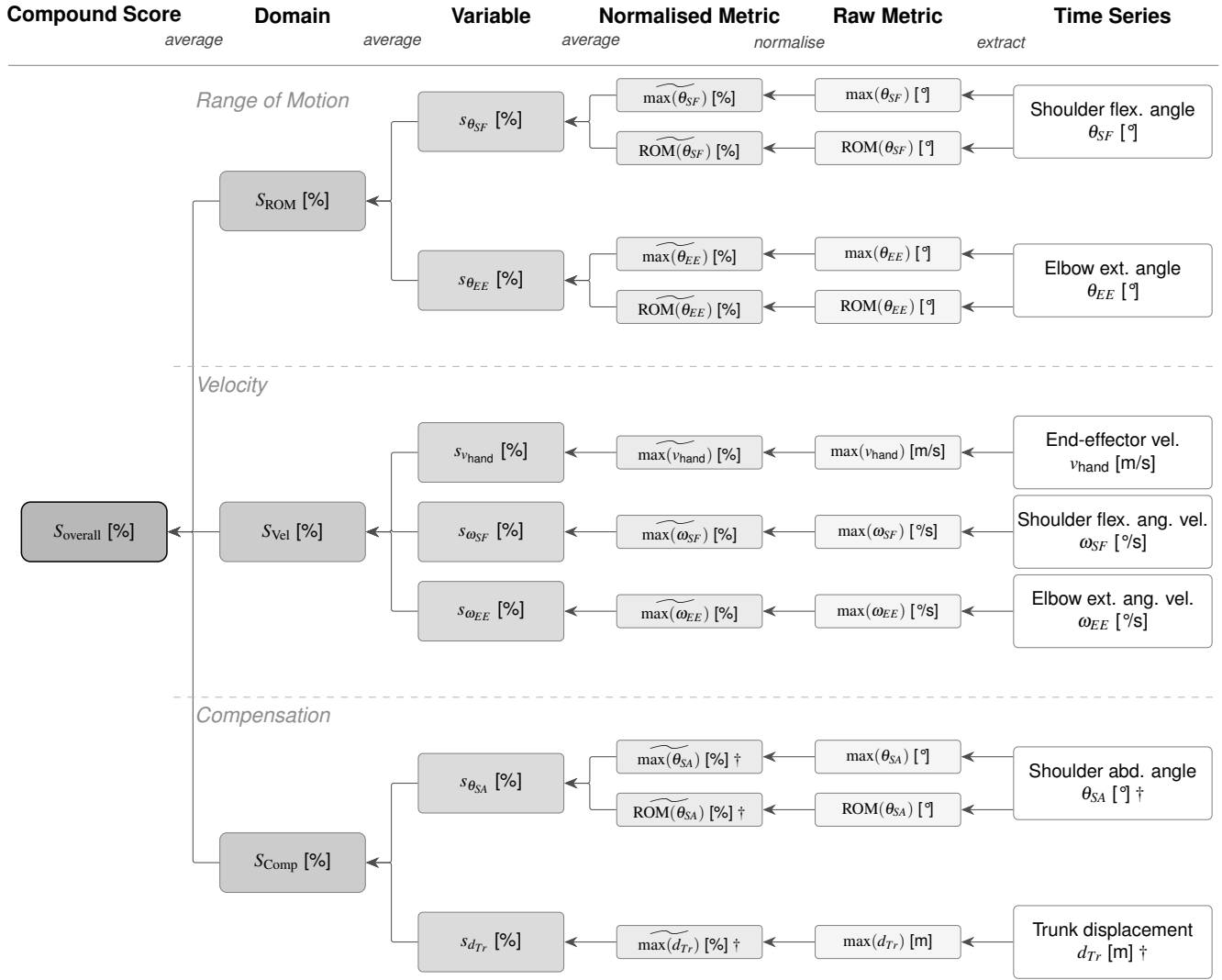
\begin{figure*}[h!]
\centering
\resizebox{\textwidth}{!}{%
\begin{tikzpicture}[
  font=\sffamily\footnotesize,
  >={Stealth[length=2mm,width=1.6mm]},
  ts/.style={draw=black!50, fill=white, rounded corners=2pt,
              minimum width=30mm, minimum height=11mm, align=center,
              inner sep=1pt},
  rm/.style={draw=black!50, fill=black!4, rounded corners=1.5pt,
              minimum width=22mm, minimum height=5mm, align=center,
              inner sep=1pt, font=\sffamily\scriptsize},
  nm/.style={draw=black!50, fill=black!8, rounded corners=1.5pt,
              minimum width=22mm, minimum height=5mm, align=center,
              inner sep=1pt, font=\sffamily\scriptsize},
  var/.style={draw=black!50, fill=black!14, rounded corners=2pt,
               minimum width=22mm, minimum height=9mm, align=center,
               inner sep=1pt},
  dom/.style={draw=black!50, fill=black!20, rounded corners=3pt,
               minimum width=22mm, minimum height=9mm, align=center,
               inner sep=1pt},
  comp/.style={draw=black, line width=0.6pt, fill=black!28, rounded corners=4pt,
                minimum width=22mm, minimum height=9mm, align=center,
                inner sep=1pt},
  arr/.style={->, thin, black!70},
  splitarr/.style={->, thin, black!70, rounded corners=1pt},
  colhead/.style={font=\sffamily\small\bfseries, align=center},
  oplabel/.style={font=\sffamily\scriptsize\itshape, align=center, text=black!75},
  banddiv/.style={dashed, black!30, thin},
  bandlabel/.style={font=\sffamily\small\itshape, text=black!45, align=left},
]

\node[ts] (ts1) at (15.9, 0)     {Shoulder flex. angle\\$\theta_{SF}$ [\textdegree]};
\node[ts] (ts2) at (15.9,-2.4)   {Elbow ext. angle\\$\theta_{EE}$ [\textdegree]};
\node[ts] (ts3) at (15.9,-5.0)   {End-effector vel.\\$v_{\text{hand}}$ [m/s]};
\node[ts] (ts4) at (15.9,-6.2)   {Shoulder flex. ang. vel.\\$\omega_{SF}$ [\textdegree/s]};
\node[ts] (ts5) at (15.9,-7.4)   {Elbow ext. ang. vel.\\$\omega_{EE}$ [\textdegree/s]};
\node[ts] (ts6) at (15.9,-10.0)  {Shoulder abd. angle\\$\theta_{SA}$ [\textdegree] $\dagger$};
\node[ts] (ts7) at (15.9,-12.4)  {Trunk displacement\\$d_{Tr}$ [m] $\dagger$};

\node[rm] (m1a) at (12.5, 0.4)   {$\max(\theta_{SF})$ [\textdegree]};
\node[rm] (m1b) at (12.5,-0.4)   {$\mathrm{ROM}(\theta_{SF})$ [\textdegree]};
\node[rm] (m2a) at (12.5,-2.0)   {$\max(\theta_{EE})$ [\textdegree]};
\node[rm] (m2b) at (12.5,-2.8)   {$\mathrm{ROM}(\theta_{EE})$ [\textdegree]};
\node[rm] (m3)  at (12.5,-5.0)   {$\max(v_{\text{hand}})$ [m/s]};
\node[rm] (m4)  at (12.5,-6.2)   {$\max(\omega_{SF})$ [\textdegree/s]};
\node[rm] (m5)  at (12.5,-7.4)   {$\max(\omega_{EE})$ [\textdegree/s]};
\node[rm] (m6a) at (12.5,-9.6)   {$\max(\theta_{SA})$ [\textdegree]};
\node[rm] (m6b) at (12.5,-10.4)  {$\mathrm{ROM}(\theta_{SA})$ [\textdegree]};
\node[rm] (m7)  at (12.5,-12.4)  {$\max(d_{Tr})$ [m]};

\draw[arr] (ts1.west |- m1a) -- (m1a.east);
\draw[arr] (ts1.west |- m1b) -- (m1b.east);
\draw[arr] (ts2.west |- m2a) -- (m2a.east);
\draw[arr] (ts2.west |- m2b) -- (m2b.east);
\draw[arr] (ts3.west) -- (m3.east);
\draw[arr] (ts4.west) -- (m4.east);
\draw[arr] (ts5.west) -- (m5.east);
\draw[arr] (ts6.west |- m6a) -- (m6a.east);
\draw[arr] (ts6.west |- m6b) -- (m6b.east);
\draw[arr] (ts7.west) -- (m7.east);

\node[nm] (n1a) at (9.5, 0.4)   {$\widetilde{\max(\theta_{SF})}$ [\%]};
\node[nm] (n1b) at (9.5,-0.4)   {$\widetilde{\mathrm{ROM}(\theta_{SF})}$ [\%]};
\node[nm] (n2a) at (9.5,-2.0)   {$\widetilde{\max(\theta_{EE})}$ [\%]};
\node[nm] (n2b) at (9.5,-2.8)   {$\widetilde{\mathrm{ROM}(\theta_{EE})}$ [\%]};
\node[nm] (n3)  at (9.5,-5.0)   {$\widetilde{\max(v_{\text{hand}})}$ [\%]};
\node[nm] (n4)  at (9.5,-6.2)   {$\widetilde{\max(\omega_{SF})}$ [\%]};
\node[nm] (n5)  at (9.5,-7.4)   {$\widetilde{\max(\omega_{EE})}$ [\%]};
\node[nm] (n6a) at (9.5,-9.6)   {$\widetilde{\max(\theta_{SA})}$ [\%] $\dagger$};
\node[nm] (n6b) at (9.5,-10.4)  {$\widetilde{\mathrm{ROM}(\theta_{SA})}$ [\%] $\dagger$};
\node[nm] (n7)  at (9.5,-12.4)  {$\widetilde{\max(d_{Tr})}$ [\%] $\dagger$};

\draw[arr] (m1a.west) -- (n1a.east);
\draw[arr] (m1b.west) -- (n1b.east);
\draw[arr] (m2a.west) -- (n2a.east);
\draw[arr] (m2b.west) -- (n2b.east);
\draw[arr] (m3.west)  -- (n3.east);
\draw[arr] (m4.west)  -- (n4.east);
\draw[arr] (m5.west)  -- (n5.east);
\draw[arr] (m6a.west) -- (n6a.east);
\draw[arr] (m6b.west) -- (n6b.east);
\draw[arr] (m7.west)  -- (n7.east);

\node[var] (v1) at (6.5, 0)      {$s_{\theta_{SF}}$ [\%]};
\node[var] (v2) at (6.5,-2.4)    {$s_{\theta_{EE}}$ [\%]};
\node[var] (v3) at (6.5,-5.0)    {$s_{v_{\text{hand}}}$ [\%]};
\node[var] (v4) at (6.5,-6.2)    {$s_{\omega_{SF}}$ [\%]};
\node[var] (v5) at (6.5,-7.4)    {$s_{\omega_{EE}}$ [\%]};
\node[var] (v6) at (6.5,-10.0)   {$s_{\theta_{SA}}$ [\%]};
\node[var] (v7) at (6.5,-12.4)   {$s_{d_{Tr}}$ [\%]};

\draw[splitarr] (n1a.west) -- ++(-0.4,0) |- (v1.east);
\draw[splitarr] (n1b.west) -- ++(-0.4,0) |- (v1.east);
\draw[splitarr] (n2a.west) -- ++(-0.4,0) |- (v2.east);
\draw[splitarr] (n2b.west) -- ++(-0.4,0) |- (v2.east);
\draw[arr] (n3.west) -- (v3.east);
\draw[arr] (n4.west) -- (v4.east);
\draw[arr] (n5.west) -- (v5.east);
\draw[splitarr] (n6a.west) -- ++(-0.4,0) |- (v6.east);
\draw[splitarr] (n6b.west) -- ++(-0.4,0) |- (v6.east);
\draw[arr] (n7.west) -- (v7.east);

\node[dom] (dROM)  at (3.5,-1.2)   {$S_{\mathrm{ROM}}$ [\%]};
\node[dom] (dVel)  at (3.5,-6.2)   {$S_{\mathrm{Vel}}$ [\%]};
\node[dom] (dComp) at (3.5,-11.2)  {$S_{\mathrm{Comp}}$ [\%]};

\draw[splitarr] (v1.west) -- ++(-0.4,0) |- (dROM.east);
\draw[splitarr] (v2.west) -- ++(-0.4,0) |- (dROM.east);
\draw[splitarr] (v3.west) -- ++(-0.4,0) |- (dVel.east);
\draw[splitarr] (v4.west) -- ++(-0.4,0) |- (dVel.east);
\draw[splitarr] (v5.west) -- ++(-0.4,0) |- (dVel.east);
\draw[splitarr] (v6.west) -- ++(-0.4,0) |- (dComp.east);
\draw[splitarr] (v7.west) -- ++(-0.4,0) |- (dComp.east);

\node[comp] (Comp) at (0.5,-6.2)  {$S_{\mathrm{overall}}$ [\%]};

\draw[splitarr] (dROM.west)  -- ++(-0.4,0) |- (Comp.east);
\draw[arr]      (dVel.west)  -- (Comp.east);
\draw[splitarr] (dComp.west) -- ++(-0.4,0) |- (Comp.east);

\node[colhead] at (0.5,  1.60) {Compound Score};
\node[colhead] at (3.5,  1.60) {Domain};
\node[colhead] at (6.5,  1.60) {Variable};
\node[colhead] at (9.5,  1.60) {Normalised Metric};
\node[colhead] at (12.5, 1.60) {Raw Metric};
\node[colhead] at (15.9, 1.60) {Time Series};

\node[oplabel] at (2.0,  1.20) {average};
\node[oplabel] at (5.0,  1.20) {average};
\node[oplabel] at (8.0,  1.20) {average};
\node[oplabel] at (11.0, 1.20) {normalise};
\node[oplabel] at (14.2, 1.20) {extract};

\draw[black!50, thin] (-0.8,0.85) -- (17.5,0.85);

\draw[banddiv] (1.7,-3.7) -- (17.5,-3.7);
\draw[banddiv] (1.7,-8.7) -- (17.5,-8.7);

\node[bandlabel] at (1.9, 0.3) [anchor=west] {Range of Motion};
\node[bandlabel] at (1.9,-4.0) [anchor=west] {Velocity};
\node[bandlabel] at (1.9,-9.0) [anchor=west] {Compensation};

\end{tikzpicture}%
}

\caption{Hierarchical construction of compound scores from kinematic metrics.
Reading right to left: seven kinematic time series (right) are summarised by
ten raw metrics (max or range of motion, ROM). Each raw metric is normalised
to a Q75-anchored percentage scale on which $100\,\%$ corresponds to the
per-metric 75th percentile (Q75) of Score~3 (clinically normal) cohort
sessions and $0\,\%$ to the per-task cohort minimum, yielding a normalised
metric $\widetilde{m}$; values are not clipped and may exceed $100\,\%$ to
indicate performance beyond the normal-performance threshold. Metrics marked with $\dagger$ are
compensation metrics for which higher raw values indicate worse function;
for these the normalisation is inverted (Q25 used as the $100\,\%$ anchor
instead of Q75) so that higher normalised values consistently reflect better
function across all metrics. Normalised metrics are averaged within each
time series to yield seven variable scores; variable scores are averaged
within each of three domains (Range of Motion, Velocity, Compensation); and
the three domain scores are averaged to yield the overall compound score.
All aggregated scores are expressed on the same Q75-anchored percentage
scale. Background shading darkens with increasing aggregation, from raw
inputs (white) to the overall compound score.
Symbols: $\theta$ joint angle; $\omega$ joint angular velocity; $v$ linear
velocity; $d$ linear displacement; subscripts denote joint
(SF: shoulder flexion; EE: elbow extension; SA: shoulder abduction;
Tr: trunk).}
\label{fig:compound_score_aggregation}
\end{figure*}

Ten kinematic metrics were extracted at the trial level. The 
guiding principle was to retain metrics that are robustly 
computable from a single trial without requiring movement phase 
segmentation, yielding stable estimates that characterise upper 
limb movement directly from the kinematic time series. This 
trial-level approach avoids any additional error introduced by 
automated phase classification, which would propagate uncertainty 
into every downstream metric. Metrics were derived from the 
kinematic time series most consistently implicated in upper limb 
movement characterisation in the literature: end-effector velocity, 
elbow angle and angular velocity, shoulder flexion angle and angular 
velocity, shoulder abduction, and trunk displacement 
\cite{alt_murphy_kinematic_2018, unger_differentiable_2025}. From 
these underlying time series, stable trial-level summary metrics 
were computed as peak values and ranges of motion, capturing the 
magnitude and extent of movement during each trial. Smoothness 
metrics were excluded as they are more susceptible to kinematic 
jitter introduced by the reduced camera count in the clinical 
environment relative to the laboratory setup. All metrics are 
listed in Figure~\ref{fig:compound_score_aggregation} (raw metrics).

\subsubsection*{Task grouping}

The analysis covered all 19 ARAT tasks, performed bilaterally 
yielding 38 task instances. As it is good clinical practice to not 
evaluate a movement based on a single repetition 
\cite{frykberg_how_2021}, kinematic metrics and clinical scores of 
the individual tasks were averaged per ARAT subgroup wherever they 
were sharing the same gross movement pattern. This resulted in 7 
analysis groups (see Table~\ref{tab:task_grouping}). A secondary 
benefit of this averaging is that the per-analysis-group clinical score 
becomes a continuous quantity on $[0, 3]$ rather than the discrete 
ordinal values 0--3 of the individual tasks, providing finer 
resolution that aligns more naturally with the continuous nature of 
the kinematic compound scores in subsequent analyses.

\subsubsection*{Compound Metrics}
To support clinical interpretability, kinematic metrics were aggregated
into compound scores summarizing movement across three domains
(Range-of-Motion, Velocity, Compensation; see
Figure~\ref{fig:compound_score_aggregation}). Referencing movement
quality to a normal motor-performance benchmark is an established way to
interpret upper-limb kinematic metrics
\cite{alt_murphy_kinematic_2018}.
Lacking a separate able-bodied cohort, we approximated this reference
with Score~3 (clinically normal) sessions. Because Score~3 carries the
scoring subjectivity this study documents, we anchored the normal
threshold at the 75th percentile (Q75) of Score~3 rather than the
median, a more conservative choice robust to that subjectivity.

Compound scores are expressed on this anchored scale: 100\,\%
corresponds to the per-metric Q75 of Score~3 sessions and 0\,\% to the
per-task cohort minimum. Values are not clipped and may exceed 100\,\%,
indicating performance beyond the normal threshold. For metrics where
higher values indicate worse performance (trunk displacement, shoulder
abduction), Q25 replaces Q75 as the 100\,\% anchor, so higher scores
always indicate better function. Where one kinematic variable is
represented by multiple metrics (e.g.\ peak and ROM of shoulder
flexion), those metrics were averaged into a single variable score so
that each variable contributes equally within its domain (see Figure \ref{fig:compound_score_aggregation}).

Domain scores were computed as the unweighted mean of their constituent
variable scores: $S_{\mathrm{ROM}}$ (shoulder flexion, elbow extension),
$S_{\mathrm{Vel}}$ (end-effector, shoulder flexion, elbow extension), and
$S_{\mathrm{Comp}}$ (trunk displacement, shoulder abduction). The overall
score ($S_{\mathrm{overall}}$) is the unweighted mean of the three domain
scores. Compound-score \emph{levels} are reported in percent (\%) while
\emph{changes} and the MCID threshold are reported in percentage points
(pp).

\subsection*{Data Analysis}

\subsubsection*{Construct Validity at the Population Level (H2)}
\label{sec:population_analysis}

Construct validity at the population level was evaluated through a 
two-tier known-groups discrimination analysis \cite{de_vet_measurement_2011} applied independently 
to each ARAT analysis group (Table~\ref{tab:task_grouping}). 
\textit{Tier~1} compared Score~0/1 (the task could not be completed) 
versus Score~2/3 (the task was completed) trials, representing the 
clinically robust boundary where strong discrimination was 
expected. \textit{Tier~2} compared Score~2 (completed with 
difficulty or abnormal time) versus Score~3 (completed normally) 
trials, probing the more subjective upper boundary of the ARAT 
scale where inter-rater variability is greatest and where 
attenuated discrimination would itself support construct validity 
by indicating that the metrics track the underlying functional 
construct rather than the ordinal, clinical scale. Group differences 
for each metric were evaluated using Mann--Whitney $U$ tests with 
rank-biserial correlation $r$ as the effect size measure, and 
significance thresholds at $p < .05$, $p < .01$, and $p < .001$. 
Discriminative capacity was further quantified using logistic 
regression with 5-fold stratified cross-validation, with 
performance reported as AUC. To assess the relative contribution 
of different movement domains, four feature sets were compared 
independently: ROM-only (four metrics), Velocity-only (three 
metrics), Compensation-only (three metrics), and all ten metrics 
combined. H2 was considered supported if Tier~1 combined-feature 
AUC met the conventional ``excellent discrimination'' threshold of 
0.85 \cite{hosmer_applied_2013} in the majority of analysis groups, 
and Tier~2 combined-feature AUC remained above chance level (AUC $> 0.5$) 
while consistently lower than Tier~1 across analysis groups.

\subsubsection*{Longitudinal Case Studies (H3)}
\label{sec:longitudinal_methods}
Two participants with the most recorded sessions were selected to 
illustrate the longitudinal behaviour of the kinematic metrics, 
examined in detail on the Grasp analysis group: P10, followed over four 
sessions all below the clinical ARAT ceiling (sub-ceiling window), and 
P20, followed over nine sessions across approximately four months, 
reaching the clinical ceiling at session~3 and remaining there 
afterwards (sub-ceiling window: sessions~1--3; post-ceiling window: 
sessions~3--9). For each participant, the per-analysis-group clinical 
ARAT score and the kinematic compound scores ($S_{\mathrm{overall}}$, 
$S_{\mathrm{ROM}}$, $S_{\mathrm{Vel}}$, $S_{\mathrm{Comp}}$) were 
plotted across sessions alongside the individual metric trajectories. 
Given the few sessions per sub-ceiling window (3--4), inferential trend 
tests are not meaningful at this $n$; we therefore characterised 
longitudinal agreement descriptively, benchmarking against an adopted 
MCID (15\,pp on the compound scale) translated from consensus 
recommendations for kinematic measurement of upper limb quality after 
stroke \cite{kwakkel_standardized_2019}. As this threshold is adopted 
rather than estimated for the present measure and cohort, it is an 
approximate benchmark; no test--retest data were available to estimate 
a measurement-error-based threshold directly.

\textit{Sub-ceiling agreement and specificity (H3a, H3b).} For each analysis group, the 
mean clinical ARAT score was rescaled to a common 0--100 axis across 
the sub-ceiling window, and the kinematic compound scores (already 
expressed on the normalised scale defined above) were used directly. 
Per-session change vectors for the clinical and each kinematic 
trajectory were then computed on this common percentage-point axis. 
The mean absolute deviation (MAD) between the clinical change vector 
and each kinematic change vector was reported as the index of 
longitudinal agreement, with smaller MAD indicating closer tracking 
of the clinical trajectory and larger MAD indicating divergence either 
in magnitude or direction. The $S_{\mathrm{overall}}$ MAD was used as 
the validity criterion, and domain compound MADs were additionally 
reported to illustrate patient-specific recovery profiles where 
present. H3a was evaluated per analysis group: the 
$S_{\mathrm{overall}}$ MAD was compared against the MCID (15\,pp) 
separately for each analysis group, with analysis groups falling below 
MCID establishing longitudinal construct validity and exceedances 
expected to coincide with the metric-task mismatches identified 
cross-sectionally (H2). Where patient-specific recovery profiles were 
present, domain MAD divergence then illustrated the added clinical 
value the validated compound makes interpretable (H3b).

\textit{Post-ceiling responsiveness (H3c).} For P20, the 
post-ceiling window (sessions in which the clinical ARAT score 
had reached and remained at the maximum of 3) was analysed 
separately. Changes were computed at two levels with distinct 
referents. For raw individual kinematic metrics, $\Delta\,\%$ was 
expressed as a percentage of the cohort-wide range of that metric 
($\Delta\,\% = 100 \times (\text{last} - \text{first}) / 
\text{cohort range}$). For compound and domain scores, $\Delta$ was 
reported in percentage points (pp) on the normalised scale, 
interpretable as the gap closed toward (or, when negative, opened 
from) the per-metric normal motor-performance threshold. H3c was 
considered supported if at least one domain compound score changed by 
$\geq 15$\,pp (one MCID) within the post-ceiling window.


\section*{Results}

\subsection*{Data Set}

A total of 1,304 ARAT tasks (trials) were recorded across 47 assessment sessions from 20 
participants. Participant characteristics are summarised in 
Supplementary Table \ref{tab:demographics}; the cohort comprised 15 males and 5 females 
(mean age 71.2\,years, standard deviation (SD)\,9.9, range 55--90) with diagnoses of stroke 
($n$=10), Parkinson's disease ($n$=6), and other neurological conditions 
($n$=4). Thirteen participants contributed more than one session, with 
the two longest monitored participants followed over approximately four 
months (9 sessions) and nine weeks (4 sessions), respectively (the 
distribution of sessions per participant is shown in 
Supplementary Figure \ref{fig:sessions-per-participant}). Following 
score-aware interquartile-range (IQR)-based outlier detection, 130 trials (10.0\%) were 
excluded, leaving 1,174 trials for analysis.

\subsection*{Data Quality of Kinematic Reconstruction (H1)}

Across all analysed trials, the mean reprojection error of the 
biomechanical reconstruction was 12.5\,$\pm$\,3.0\,px (median 12.1\,px, 
IQR 10.5--14.0\,px). This falls within the acceptable 10--20\,px range 
previously established for this reconstruction algorithm in clinical 
use, modestly above the 5--10\,px laboratory standard 
\cite{unger_differentiable_2025} as expected given the reduced 
three-camera configuration and more frequent occlusions inherent to a 
clinical setting. The adequacy criterion for H1 is therefore met.

\begin{figure}[hb]
    \centering
    \begin{minipage}[t]{0.60\textwidth}
        \vspace{0pt}
        \captionsetup{type=table,width=\linewidth,justification=justified,singlelinecheck=false,aboveskip=0pt,belowskip=8pt}
        \caption{Reprojection error of MMC by ARAT score group over all 1304 trials (pre outlier removal). Group-level descriptives (median, interquartile range), Kruskal--Wallis omnibus test, and pairwise Dunn's post-hoc tests with Holm correction. $r$ denotes the rank-biserial correlation as the pairwise effect size. Effect-size labels follow conventional thresholds: $\eta^2_H<0.01$ negligible, $0.01$--$0.06$ small, $0.06$--$0.14$ medium, $>0.14$ large; $|r|<0.10$ negligible, $0.10$--$0.30$ small, $0.30$--$0.50$ medium, $\geq 0.50$ large. Significance: $^{*}p<.05$, $^{**}p<.01$, $^{***}p<.001$, n.s.\ = not significant.}
        \label{tab:reprojection_by_score}
        \footnotesize
        \setlength{\tabcolsep}{2pt}
        \renewcommand{\arraystretch}{0.88}
        \renewcommand{\arraystretch}{1.3}
\setlength{\tabcolsep}{3pt}
\begin{tabular}{@{} c @{\hspace{0.4cm}} c @{}}
\hline\hline
\begin{tabular}{ c c c c c c }
\multicolumn{6}{c}{\textit{Group descriptives}} \\
\hline
Score & $n$ & Median & IQR        & Mean  & SD    \\
      &     & (px)   & (px)       & (px)  & (px)  \\
\hline
0 & 28 & 13.90 & [10.73, 16.27] & 14.04 & 3.97 \\
1 & 113 & 11.36 & [10.05, 14.32] & 12.16 & 3.32 \\
2 & 650 & 11.84 & [10.47, 13.79] & 12.47 & 3.00 \\
3 & 513 & 12.32 & [10.74, 13.96] & 12.61 & 2.82 \\
\hline
\multicolumn{6}{c}{\textit{Kruskal--Wallis}: $H(3){=}9.85$, $p{=}0.020$} \\
\multicolumn{6}{c}{$\eta^2_H{=}0.0053$ (negligible)} \\
\end{tabular}
&
\begin{tabular}{ c c c c c }
\multicolumn{5}{c}{\textit{Pairwise (Dunn's, Holm-corrected)}} \\
\hline
Pair & $\Delta_{\mathrm{med}}$ & $r$ & $|r|$ & $p_{\mathrm{Holm}}$ \\
     & (px)                    &     &       &                    \\
\hline
0 vs 1 & $-2.55$ & $-0.257$ & small & 0.059 \\
0 vs 2 & $-2.07$ & $-0.249$ & small & 0.096 \\
0 vs 3 & $-1.58$ & $-0.229$ & small & 0.236 \\
1 vs 2 & $+0.48$ & $+0.061$ & negl. & 0.365 \\
1 vs 3 & $+0.97$ & $+0.107$ & small & 0.236 \\
2 vs 3 & $+0.49$ & $+0.062$ & negl. & 0.236 \\
\end{tabular}
\\
\hline\hline
\end{tabular}

    \end{minipage}\hfill
    \begin{minipage}[t]{0.36\textwidth}
        \vspace{0pt}
        \centering
        \includegraphics[width=\textwidth]{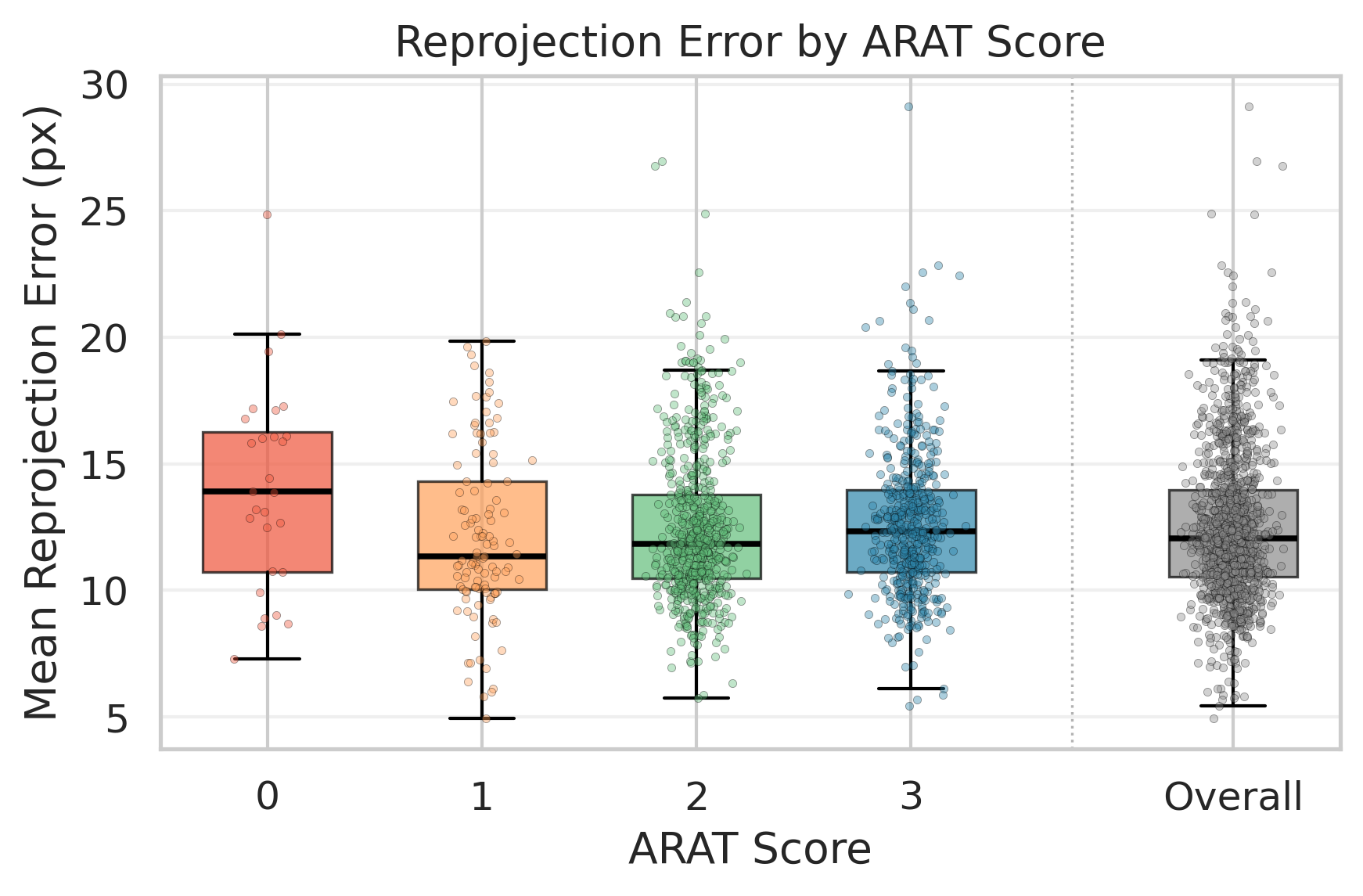}
        \caption{Mean per-trial reprojection error (pixels) by clinical ARAT score (0--3) and pooled across all 1304 trials (right). Largely overlapping distributions confirm that reconstruction quality does not systematically depend on patient impairment level ($\eta^2_H{=}0.0053{<}0.06$; Table~\ref{tab:reprojection_by_score}).}
        \label{fig:reprojection-error-by-score}

    \end{minipage}
\end{figure}

Reprojection error was largely stable across the four ARAT score 
groups (all trials pooled by their clinical ARAT score, 0--3, each trial 
contributing one mean reprojection-error value) 
(Table~\ref{tab:reprojection_by_score}, 
Figure~\ref{fig:reprojection-error-by-score}). Median reprojection 
error differed by at most 2.5\,px across the four score groups 
(13.9\,px for Score~0 vs.\ 11.4\,px for Score~1, the largest pairwise 
gap), a small difference relative to the overall error distribution 
(IQR 3.5\,px across all trials). A Kruskal--Wallis test indicated a 
statistically significant difference across score groups 
($H(3) = 9.85$, $p = .020$), but the associated effect size was 
negligible ($\eta^2_H = 0.0053$), indicating that score group explained 
essentially none of the variance in reconstruction quality and 
satisfying the robustness criterion ($\eta^2_H < 0.06$). Pairwise 
Dunn's tests with Holm correction revealed no significant differences 
between any score groups after correction (all $p_{\mathrm{Holm}} > .05$), 
with effect sizes ranging from negligible to small ($|r| \leq 0.26$). 
The largest, albeit non-significant, difference involved Score~0 
trials, which showed slightly higher median error (13.9\,px) than 
Scores~1--3 (11.4--12.3\,px), consistent with the greater movement 
variability and more frequent occlusions expected in the most impaired 
performances.

With both adequacy (12.5\,px within the 10--20\,px acceptable range) 
and robustness ($\eta^2_H = 0.0053 < 0.06$) criteria met, H1 is 
supported: adequate and uniform kinematic reconstruction is achievable 
across the full range of clinical ability in a routine, reduced-camera 
clinical setting.

\subsection*{Construct Validity at the Population Level (H2)}
We assessed construct validity through two-tier known-groups 
discrimination. Tier~1 (Score~0/1 vs.\ Score~2/3) probes the robust 
boundary between completed and uncompleted tasks, where strong 
discrimination is expected; Tier~2 (Score~2 vs.\ Score~3) probes the 
more subjective boundary between completions of differing quality, 
where discrimination lower than Tier~1 would itself support validity.

\subsubsection*{Tier~1: Score~0/1 vs.\ Score~2/3 Discrimination}

\begin{figure}[t]
    \centering
    \includegraphics[width=\linewidth]{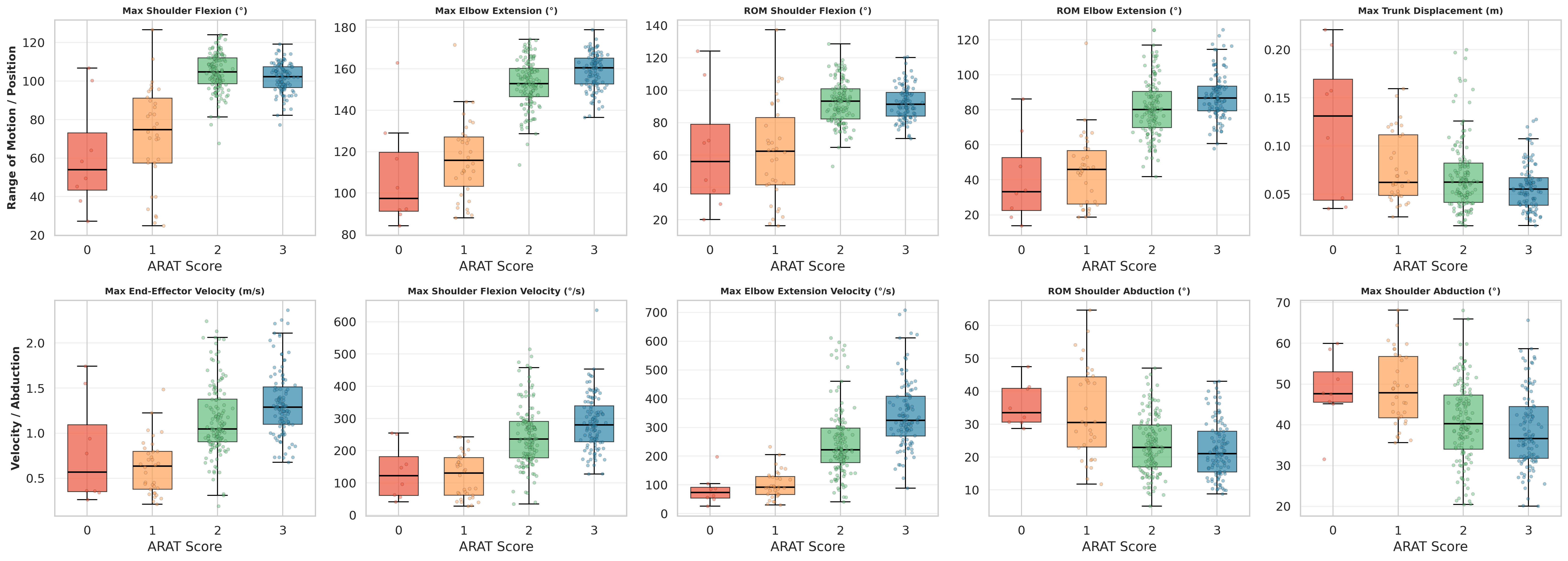}
    \caption{Distribution of kinematic metrics across clinical ARAT 
    scores (0--3) for all trials in the Grasp analysis group. Individual trials 
    shown as jittered points. Outliers removed (IQR $\times$ 3.0).}
    \label{fig:grasp_boxplots_by_score}
\end{figure}

Tier~1 discrimination between Score~0/1 and Score~2/3 trials was 
strong across six of seven analysis groups, with combined-feature AUC 
ranging from 0.91 to 1.00 
(Table~\ref{tab:results_all_tasks_merged}; 
Figure~\ref{fig:grasp_boxplots_by_score} illustrates the underlying 
metric distributions for all trials in the Grasp analysis group), demonstrating 
that the MMC-derived kinematic metrics reliably capture the 
fundamental distinction between task completion and failure to 
complete.

For pick-and-place tasks (Grasp, Grip, Pinch), combined-feature 
discrimination was near-perfect (AUC 0.96--$>$0.99, see Table \ref{tab:results_all_tasks_merged}). The most 
consistent individual discriminators were elbow extension metrics (both ROM $r = 0.77$--$0.93$ and peak velocity $r = 0.82$--$0.93$), reflecting the requirement to achieve near-full arm extension to 
transport the object to the target. Score~2/3 trials also showed 
substantially higher end-effector and joint velocities than Score~0/1 
trials (see Supplementary Table \ref{tab:descriptive_stats_merged}). ROM metrics 
discriminated strongly on their own (ROM-only AUC $\geq 0.95$), with 
velocity-only somewhat lower (0.90--0.94); combining the domains 
yielded the highest discrimination in every case (AUC $\geq 0.96$).

\begin{figure}[h!]
    \centering

    \begin{subfigure}{\linewidth}
        \centering
        \includegraphics[width=\linewidth]{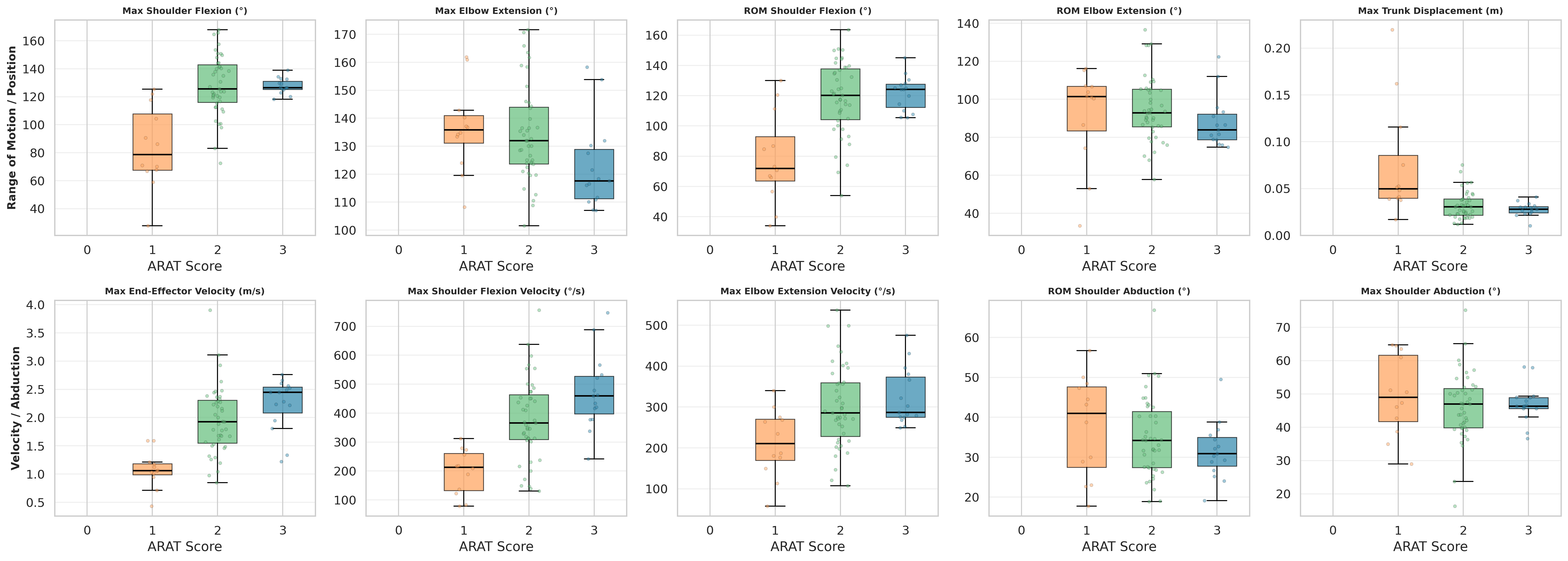}
        \caption{Hand behind head}
        \label{fig:hand_behind_head_boxplots_by_score}
    \end{subfigure}

    \begin{subfigure}{\linewidth}
        \centering
        \includegraphics[width=\linewidth]{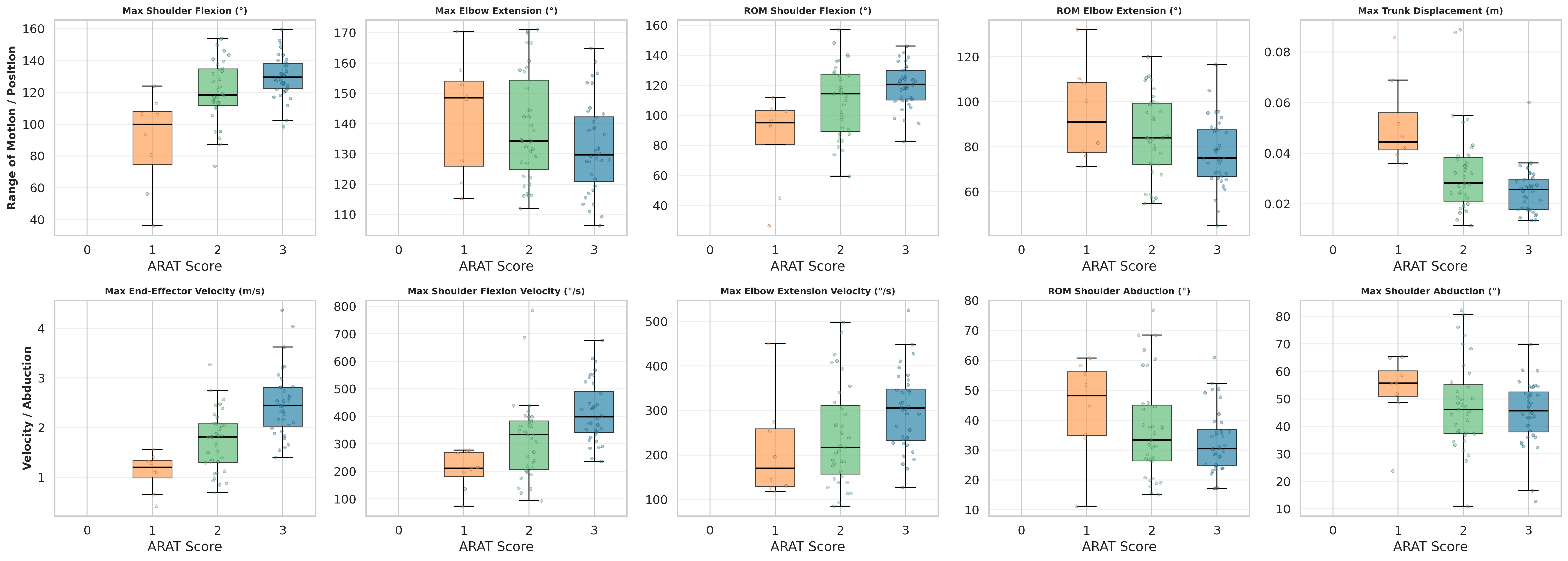}
        \caption{Hand on top of head}
        \label{fig:hand_on_top_of_head_boxplots_by_score}
    \end{subfigure}

    \begin{subfigure}{\linewidth}
        \centering
        \includegraphics[width=\linewidth]{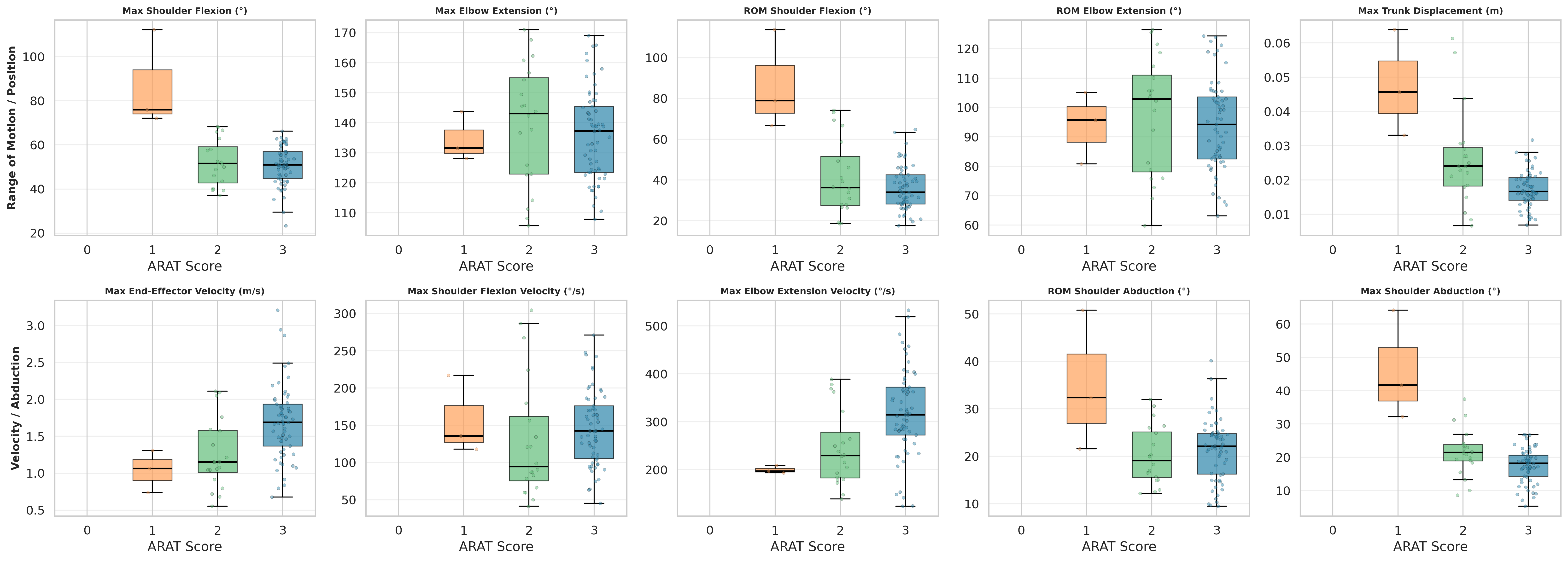}
        \caption{Hand to mouth}
        \label{fig:hand_to_mouth_boxplots_by_score}
    \end{subfigure}

    \caption{Distribution of kinematic metrics across clinical ARAT scores (0--3) for gross movement tasks per analysis group. Individual trials shown as jittered points. Outliers removed (IQR $\times$ 3.0).}
    \label{fig:population_adl_boxplots_by_score}
\end{figure}

Gross motor tasks (Hand Behind Head, Hand on Top of Head, Hand to 
Mouth) similarly achieved high combined-feature discrimination 
(AUC 0.91--1.00; per-group metric distributions are shown in 
Supplementary Figure \ref{fig:population_boxplots_by_score}, \ref{fig:population_adl_boxplots_by_score}), with the dominant single domain varying across 
tasks (ROM for Hand to Mouth, velocity for Hand Behind Head, 
compensation for Hand on Top of Head) and the combined feature set 
matching or exceeding the best single domain in every case. Hand to 
Mouth showed the strongest discrimination of all analysis groups 
(AUC = 1.00), with Score~0/1 trials characterised by
higher shoulder flexion.

Trunk displacement was a significant discriminator across all 
pick-and-place and gross motor tasks, with Score~0/1 trials showing 
greater trunk compensation. As a standalone domain, compensation metrics 
discriminated Score~0/1 from Score~2/3 trials well above chance 
across all analysis groups (compensation-only AUC 0.72--0.97); for Grip 
(compensation-only AUC 0.96) and Hand to Mouth (0.97) compensation 
nearly matched ROM, and for Hand on Top of Head it was the strongest 
single domain (0.87 vs.\ ROM 0.84, velocity 0.86), indicating 
that compensatory trunk and shoulder recruitment alone carries 
substantial information about whether a task can be completed.

The Water Pouring task was an exception, showing worse performance than the other tasks (combined-feature AUC 0.75) which is 
attributable to metric-task mismatch rather than reconstruction 
failure: the current metric set and metrics category assignment (ROM, VEL, COMP) is mostly optimised for reach-and-transport 
movements. Such per-task construct-validity limitations bound the generalisation of H2 but do 
not affect the other analysis groups in which the metric set is 
well-matched to the task.

Together, Tier~1 discrimination exceeded the AUC $\geq 0.85$ 
``excellent discrimination'' threshold \cite{hosmer_applied_2013} in 
six of seven analysis groups (AUC 0.91--1.00), supporting H2 at the 
Score~0/1 vs.\ Score~2/3 boundary.

\subsubsection*{Tier~2: Score~2 vs.\ Score~3 Discrimination}

As anticipated, Tier~2 discrimination between Score~2 and Score~3 
trials was substantially lower than Tier~1 across all analysis groups 
(combined-feature AUC 0.70--0.84; 
Table~\ref{tab:results_all_tasks_merged}), reflecting reasonable 
overlap between the two groups. Compared to Tier~1, velocity metrics 
contributed more strongly to discrimination in Tier~2, sometimes 
achieving higher discriminative ability than ROM alone, for example, in the Grasp 
group (velocity-only: 0.75 vs.\ ROM-only: 0.70). This reflects the 
overall faster movement execution in Score~3 trials compared to 
Score~2 (see Supplementary Table \ref{tab:descriptive_stats_merged}).

For pick-and-place tasks (AUC 0.77--0.84), maximum elbow extension 
velocity showed the largest individual discriminative ability. Trunk displacement 
remained a significant discriminator for Grip and Pinch, but not for 
Grasp. Gross motor tasks showed slightly lower Tier~2 discrimination 
(AUC 0.70--0.82) compared to pick-and-place tasks, consistent with 
their lower performance in Tier~1; velocity metrics also dominated 
for gross motor tasks. In Water Pouring, velocity metrics now drove 
discrimination (AUC 0.83), reversing the pattern observed in 
Tier~1.

Velocity metrics were a more dominant domain in Tier~2 than in 
Tier~1, suggesting that once the task is completed, movement speed 
strongly differentiates performance quality. Compensation metrics were the weakest standalone domain at the 
Score~2 vs.\ Score~3 boundary (compensation-only AUC 0.59--0.68 across 
most tasks). This likely reflects the limited sensitivity of the 
present general-purpose compensation metrics (trunk displacement, 
shoulder abduction) to the subtler compensatory differences that 
distinguish good from normal completions, rather than an absence of 
compensatory information at this boundary \cite{unger_using_2025, sauerzopf_evaluating_2024, pohl_metrics_2026}; the per-task suitability of 
the metric set is considered further in the Discussion.
Hand to Mouth was the notable exception, where compensation remained 
the strongest single domain (AUC 0.73), driven by persistent 
shoulder abduction and trunk displacement in slower, more effortful 
completions, the one analysis group where the compensatory metrics 
distinguish Score~2 from Score~3 completions.

Tier~2 discrimination (AUC 0.70--0.84) remained above chance in all 
seven analysis groups and reached the conventional ``acceptable 
discrimination'' threshold of 0.7 \cite{hosmer_applied_2013} throughout, 
while remaining consistently lower than Tier~1 across the 
full set, the differential pattern expected of a valid construct 
measure tracking the underlying function rather than the ordinal 
scale. Combined, the Tier~1 and Tier~2 results support H2.

\begin{table}[!tb]
\caption{Known-groups construct validity across all seven task groups. Top: pick-and-place tasks (Grasp, Grip, Pinch); bottom: gross-motor and pouring tasks (Water Pouring, Hand Behind Head, Hand on Top of Head, Hand to Mouth). Rank-biserial $r$ from Mann--Whitney $U$ tests. LR AUC = logistic regression area under the ROC curve using the feature subset indicated. $^{*}p<.05$, $^{**}p<.01$, $^{***}p<.001$.}
\label{tab:results_all_tasks_merged}
\begin{center}
\textit{Pick-and-place tasks} \\[2pt]
\begin{tabular*}{\textwidth}{ @{\extracolsep{\fill}} l rr rr rr }
\hline\hline
 & \multicolumn{2}{c}{Grasp} & \multicolumn{2}{c}{Grip} & \multicolumn{2}{c}{Pinch} \\
\cmidrule(lr){2-3}\cmidrule(lr){4-5}\cmidrule(lr){6-7}
Metric & \multicolumn{1}{c}{Tier 1} & \multicolumn{1}{c}{Tier 2} & \multicolumn{1}{c}{Tier 1} & \multicolumn{1}{c}{Tier 2} & \multicolumn{1}{c}{Tier 1} & \multicolumn{1}{c}{Tier 2} \\
 & \multicolumn{1}{c}{(42 vs 248)} & \multicolumn{1}{c}{(140 vs 108)} & \multicolumn{1}{c}{(12 vs 249)} & \multicolumn{1}{c}{(110 vs 139)} & \multicolumn{1}{c}{(44 vs 271)} & \multicolumn{1}{c}{(194 vs 77)} \\
\hline
$\max(\theta_{SF})$ [$^\circ$] & 0.79$^{***}$ & -0.20$^{**}$\phantom{$^{*}$} & 0.62$^{***}$ & -0.31$^{***}$ & 0.73$^{***}$ & -0.34$^{***}$ \\
$\max(\theta_{EE})$ [$^\circ$] & 0.89$^{***}$ & 0.33$^{***}$ & 0.95$^{***}$ & 0.25$^{***}$ & 0.93$^{***}$ & -0.01\phantom{$^{***}$} \\
$\mathrm{ROM}(\theta_{SF})$ [$^\circ$] & 0.61$^{***}$ & -0.03\phantom{$^{***}$} & 0.38$^{*}$\phantom{$^{**}$} & -0.02\phantom{$^{***}$} & 0.44$^{***}$ & 0.11\phantom{$^{***}$} \\
$\mathrm{ROM}(\theta_{EE})$ [$^\circ$] & 0.88$^{***}$ & 0.28$^{***}$ & 0.93$^{***}$ & 0.15$^{*}$\phantom{$^{**}$} & 0.77$^{***}$ & 0.08\phantom{$^{***}$} \\
$\max(d_{Tr})$ [m] & -0.23$^{*}$\phantom{$^{**}$} & -0.17$^{*}$\phantom{$^{**}$} & -0.88$^{***}$ & -0.23$^{**}$\phantom{$^{*}$} & -0.62$^{***}$ & -0.31$^{***}$ \\
$\max(v_{\mathrm{hand}})$ [m/s] & 0.74$^{***}$ & 0.30$^{***}$ & 0.61$^{***}$ & 0.37$^{***}$ & 0.63$^{***}$ & 0.26$^{***}$ \\
$\max(\omega_{SF})$ [$^\circ$/s] & 0.77$^{***}$ & 0.28$^{***}$ & 0.53$^{**}$\phantom{$^{*}$} & 0.37$^{***}$ & 0.58$^{***}$ & 0.25$^{**}$\phantom{$^{*}$} \\
$\max(\omega_{EE})$ [$^\circ$/s] & 0.89$^{***}$ & 0.52$^{***}$ & 0.93$^{***}$ & 0.45$^{***}$ & 0.82$^{***}$ & 0.33$^{***}$ \\
$\mathrm{ROM}(\theta_{SA})$ [$^\circ$] & -0.50$^{***}$ & -0.10\phantom{$^{***}$} & -0.52$^{**}$\phantom{$^{*}$} & -0.14\phantom{$^{***}$} & -0.27$^{**}$\phantom{$^{*}$} & -0.04\phantom{$^{***}$} \\
$\max(\theta_{SA})$ [$^\circ$] & -0.50$^{***}$ & -0.14\phantom{$^{***}$} & -0.66$^{***}$ & -0.25$^{***}$ & -0.25$^{**}$\phantom{$^{*}$} & -0.05\phantom{$^{***}$} \\
\hline
LR AUC (vel. only) & 0.94\phantom{$^{***}$} & 0.75\phantom{$^{***}$} & 0.94\phantom{$^{***}$} & 0.72\phantom{$^{***}$} & 0.90\phantom{$^{***}$} & 0.62\phantom{$^{***}$} \\
LR AUC (ROM only) & 0.95\phantom{$^{***}$} & 0.70\phantom{$^{***}$} & 0.97\phantom{$^{***}$} & 0.77\phantom{$^{***}$} & 0.96\phantom{$^{***}$} & 0.70\phantom{$^{***}$} \\
LR AUC (comp.) & 0.76\phantom{$^{***}$} & 0.61\phantom{$^{***}$} & 0.96\phantom{$^{***}$} & 0.63\phantom{$^{***}$} & 0.82\phantom{$^{***}$} & 0.62\phantom{$^{***}$} \\
LR AUC (all) & 0.96\phantom{$^{***}$} & 0.78\phantom{$^{***}$} & $>$0.99\phantom{$^{***}$} & 0.84\phantom{$^{***}$} & 0.99\phantom{$^{***}$} & 0.77\phantom{$^{***}$} \\
\hline\hline
\end{tabular*}

\vspace{3ex}
\textit{Gross-motor and pouring tasks} \\[2pt]
\begin{tabular*}{\textwidth}{ @{\extracolsep{\fill}} l rr rr rr rr }
\hline\hline
 & \multicolumn{2}{c}{Water Pouring} & \multicolumn{2}{c}{Hand Behind Head} & \multicolumn{2}{c}{Hand on Top of Head} & \multicolumn{2}{c}{Hand to Mouth} \\
\cmidrule(lr){2-3}\cmidrule(lr){4-5}\cmidrule(lr){6-7}\cmidrule(lr){8-9}
Metric & \multicolumn{1}{c}{Tier 1} & \multicolumn{1}{c}{Tier 2} & \multicolumn{1}{c}{Tier 1} & \multicolumn{1}{c}{Tier 2} & \multicolumn{1}{c}{Tier 1} & \multicolumn{1}{c}{Tier 2} & \multicolumn{1}{c}{Tier 1} & \multicolumn{1}{c}{Tier 2} \\
 & \multicolumn{1}{c}{(5 vs 70)} & \multicolumn{1}{c}{(44 vs 26)} & \multicolumn{1}{c}{(12 vs 58)} & \multicolumn{1}{c}{(43 vs 15)} & \multicolumn{1}{c}{(8 vs 71)} & \multicolumn{1}{c}{(35 vs 36)} & \multicolumn{1}{c}{(3 vs 81)} & \multicolumn{1}{c}{(20 vs 61)} \\
\hline
$\max(\theta_{SF})$ [$^\circ$] & 0.65$^{*}$\phantom{$^{**}$} & 0.06\phantom{$^{***}$} & 0.81$^{***}$ & -0.01\phantom{$^{***}$} & 0.77$^{***}$ & 0.30$^{*}$\phantom{$^{**}$} & -1.00$^{***}$ & -0.04\phantom{$^{***}$} \\
$\max(\theta_{EE})$ [$^\circ$] & -0.19\phantom{$^{***}$} & 0.15\phantom{$^{***}$} & -0.25\phantom{$^{***}$} & -0.47$^{**}$\phantom{$^{*}$} & -0.21\phantom{$^{***}$} & -0.21\phantom{$^{***}$} & 0.04\phantom{$^{***}$} & -0.12\phantom{$^{***}$} \\
$\mathrm{ROM}(\theta_{SF})$ [$^\circ$] & 0.34\phantom{$^{***}$} & 0.13\phantom{$^{***}$} & 0.71$^{***}$ & 0.06\phantom{$^{***}$} & 0.64$^{**}$\phantom{$^{*}$} & 0.23\phantom{$^{***}$} & -0.97$^{***}$ & -0.08\phantom{$^{***}$} \\
$\mathrm{ROM}(\theta_{EE})$ [$^\circ$] & -0.42\phantom{$^{***}$} & -0.18\phantom{$^{***}$} & -0.14\phantom{$^{***}$} & -0.30\phantom{$^{***}$} & -0.36\phantom{$^{***}$} & -0.26\phantom{$^{***}$} & 0.04\phantom{$^{***}$} & -0.12\phantom{$^{***}$} \\
$\max(d_{Tr})$ [m] & -0.55$^{*}$\phantom{$^{**}$} & 0.14\phantom{$^{***}$} & -0.67$^{***}$ & -0.08\phantom{$^{***}$} & -0.81$^{***}$ & -0.29$^{*}$\phantom{$^{**}$} & -0.96$^{***}$ & -0.47$^{**}$\phantom{$^{*}$} \\
$\max(v_{\mathrm{hand}})$ [m/s] & -0.31\phantom{$^{***}$} & 0.55$^{***}$ & 0.86$^{***}$ & 0.40$^{*}$\phantom{$^{**}$} & 0.78$^{***}$ & 0.61$^{***}$ & 0.64\phantom{$^{***}$} & 0.47$^{**}$\phantom{$^{*}$} \\
$\max(\omega_{SF})$ [$^\circ$/s] & 0.03\phantom{$^{***}$} & 0.61$^{***}$ & 0.80$^{***}$ & 0.39$^{*}$\phantom{$^{**}$} & 0.72$^{***}$ & 0.46$^{***}$ & -0.18\phantom{$^{***}$} & 0.30$^{*}$\phantom{$^{**}$} \\
$\max(\omega_{EE})$ [$^\circ$/s] & -0.40\phantom{$^{***}$} & 0.24\phantom{$^{***}$} & 0.55$^{**}$\phantom{$^{*}$} & 0.15\phantom{$^{***}$} & 0.37\phantom{$^{***}$} & 0.35$^{*}$\phantom{$^{**}$} & 0.69$^{*}$\phantom{$^{**}$} & 0.47$^{**}$\phantom{$^{*}$} \\
$\mathrm{ROM}(\theta_{SA})$ [$^\circ$] & -0.66$^{*}$\phantom{$^{**}$} & -0.34$^{*}$\phantom{$^{**}$} & -0.16\phantom{$^{***}$} & -0.18\phantom{$^{***}$} & -0.38\phantom{$^{***}$} & -0.10\phantom{$^{***}$} & -0.64\phantom{$^{***}$} & 0.12\phantom{$^{***}$} \\
$\max(\theta_{SA})$ [$^\circ$] & -0.09\phantom{$^{***}$} & -0.10\phantom{$^{***}$} & -0.15\phantom{$^{***}$} & 0.04\phantom{$^{***}$} & -0.41\phantom{$^{***}$} & -0.04\phantom{$^{***}$} & -0.98$^{***}$ & -0.39$^{**}$\phantom{$^{*}$} \\
\hline
LR AUC (vel. only) & 0.69\phantom{$^{***}$} & 0.83\phantom{$^{***}$} & 0.92\phantom{$^{***}$} & 0.66\phantom{$^{***}$} & 0.86\phantom{$^{***}$} & 0.75\phantom{$^{***}$} & 0.86\phantom{$^{***}$} & 0.72\phantom{$^{***}$} \\
LR AUC (ROM only) & 0.92\phantom{$^{***}$} & 0.64\phantom{$^{***}$} & 0.91\phantom{$^{***}$} & 0.69\phantom{$^{***}$} & 0.84\phantom{$^{***}$} & 0.61\phantom{$^{***}$} & 0.98\phantom{$^{***}$} & 0.57\phantom{$^{***}$} \\
LR AUC (comp.) & 0.75\phantom{$^{***}$} & 0.60\phantom{$^{***}$} & 0.72\phantom{$^{***}$} & 0.68\phantom{$^{***}$} & 0.87\phantom{$^{***}$} & 0.59\phantom{$^{***}$} & 0.97\phantom{$^{***}$} & 0.73\phantom{$^{***}$} \\
LR AUC (all) & 0.75\phantom{$^{***}$} & 0.83\phantom{$^{***}$} & 0.97\phantom{$^{***}$} & 0.70\phantom{$^{***}$} & 0.91\phantom{$^{***}$} & 0.73\phantom{$^{***}$} & 1.00\phantom{$^{***}$} & 0.82\phantom{$^{***}$} \\
\hline\hline
\end{tabular*}
\end{center}
\end{table}

\newpage
\subsection*{Exploratory Longitudinal Case Studies (H3)}
\begin{figure} [htbp]
    \centering
    \includegraphics[width=1\linewidth]{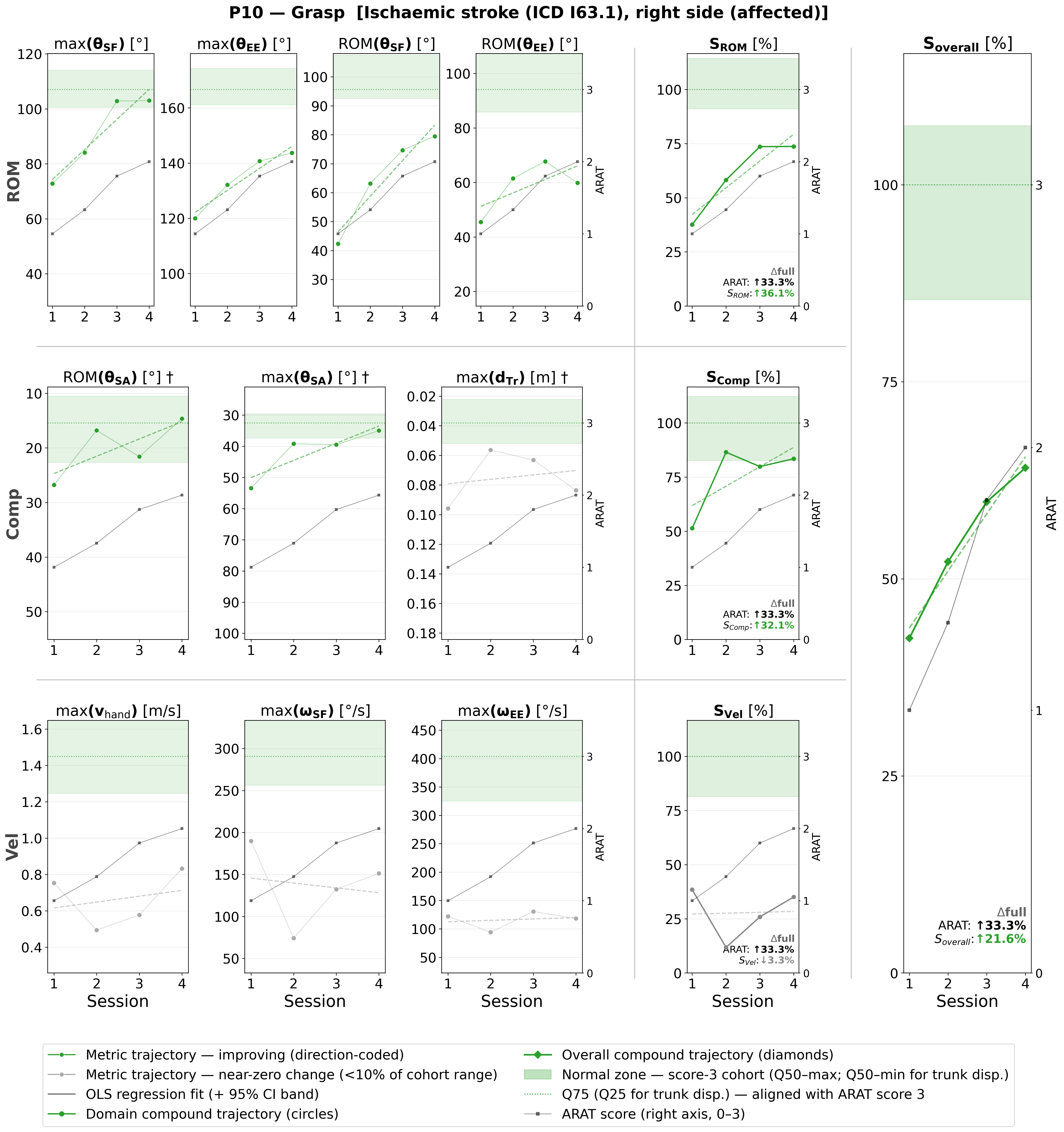}
    \caption{Longitudinal kinematic monitoring of P10's affected side 
    (ischaemic stroke) across four ARAT sessions, Grasp analysis 
    group. \textit{Panel grid}: rows are the three kinematic domains 
    (ROM, Compensation, Velocity), each with its raw metrics, a domain 
    compound score ($S_{\mathrm{ROM}}$, $S_{\mathrm{Comp}}$, 
    $S_{\mathrm{Vel}}$), and at far right the overall compound 
    ($S_{\mathrm{overall}}$). Compound panels are shown on the 
    normalised 0--100\,\% scale (100\,\% = per-metric Q75 of Score~3 
    sessions, Q25 for inverted compensation metrics marked $\dagger$; 
    see Figure~\ref{fig:compound_score_aggregation}); values are 
    unclipped and may exceed 100\,\%. Raw-metric 
    panels show absolute units. The patient's mean clinical ARAT score 
    across the Grasp analysis group tasks (0--3) is overlaid on the 
    secondary axis (black squares). 
    \textit{Reference bands}: green shading marks the normal motor-performance zone 
    (Q50--max of Score~3 sessions; Q50--min for inverted metrics); the 
    dotted line marks Q75, aligned with ARAT score~3. 
    \textit{Annotation}: each compound panel reports 
    $\Delta_{\mathrm{full}}$, the first-to-last-session change (pp), for 
    both ARAT and the compound; green $\geq 15$\,pp (MCID), red 
    $\leq -15$\,pp, grey otherwise. P10 stayed below the clinical 
    ceiling throughout, so the full window is entirely sub-ceiling.}
    \label{fig:p10_sensitivity}
\end{figure}

\begin{figure} [htbp]
    \centering
    \includegraphics[width=1\linewidth]{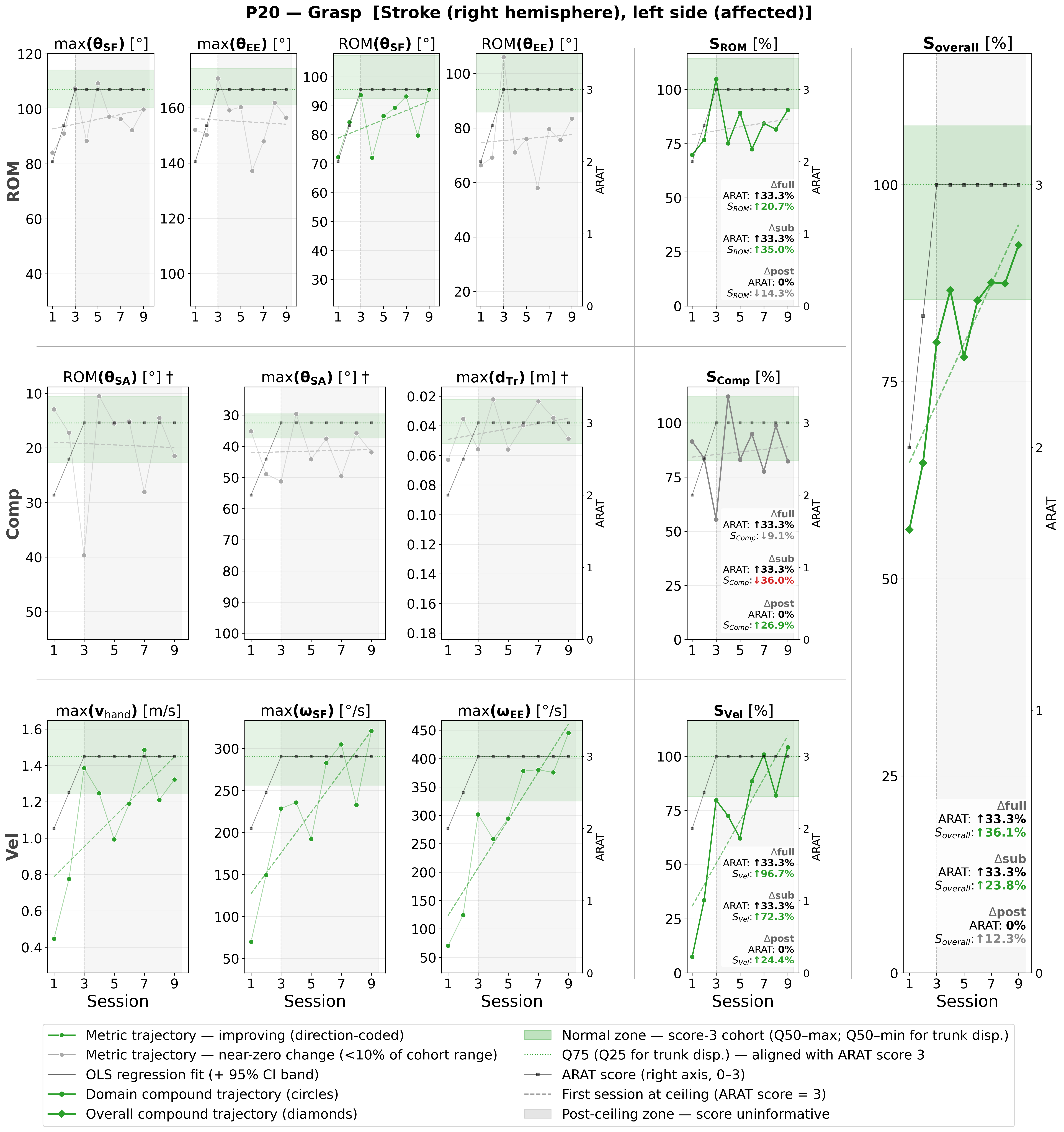}
    \caption{Longitudinal kinematic monitoring of P20's affected side 
    (stroke, right hemisphere) across nine ARAT sessions over 
    approximately four months, Grasp analysis group. Panel grid, 
    scale, reference bands, and the overlaid mean clinical ARAT score 
    as in Figure~\ref{fig:p10_sensitivity}. The vertical dashed line 
    marks the first session at which the clinical ceiling was reached 
    and held; the grey background marks the post-ceiling zone. 
    \textit{Annotation}: each compound panel reports three changes (pp) 
    for both ARAT and the compound: $\Delta_{\mathrm{full}}$ (first to 
    last), $\Delta_{\mathrm{sub}}$ (sub-ceiling, to first 
    ceiling-attaining session), and $\Delta_{\mathrm{post}}$ 
    (post-ceiling, from that session to last); by construction 
    $\Delta_{\mathrm{ARAT,\,post}} = 0$. Colouring as in 
    Figure~\ref{fig:p10_sensitivity}.}
    \label{fig:Longitudinal_P20}
\end{figure}

We illustrate longitudinal behaviour in two participants: P10 (four 
sub-ceiling sessions) and P20 (nine sessions, reaching the clinical 
ceiling at session~3), with per-session trajectories shown in 
Figures~\ref{fig:p10_sensitivity} and~\ref{fig:Longitudinal_P20}. As 
detailed in Methods, longitudinal change is reported descriptively as 
the mean absolute deviation (MAD) between clinical and kinematic 
change vectors, benchmarked against the adopted MCID (15\,pp). We 
evaluate longitudinal agreement (MAD) across the six analysis groups 
retained after excluding Water Pouring, and illustrate the 
recovery-profile contrast in detail on the Grasp analysis group.

Water Pouring is excluded from the longitudinal case-study analyses 
below for the metric-task mismatch discussed at Tier~1. 
The remaining six analysis groups (Grasp, Grip, Pinch, Hand Behind Head, Hand on Top of Head, Hand 
to Mouth) are retained.

\subsubsection*{Sub-Ceiling: Overall Concordance with Patient-Specific Domain Divergence (H3a, H3b)}

The longitudinal case studies provide two related forms of evidence. 
First, the overall kinematic compound score ($S_{\mathrm{overall}}$) 
tracked the clinical ARAT score closely across sub-ceiling sessions in 
both patients, indicating that the overall kinematic summary captures 
the same functional change as the clinical score (longitudinal 
construct validity). Second, beneath this consistent overall agreement, 
the domain compound scores (range of motion $S_{\mathrm{ROM}}$, velocity $S_{\mathrm{Vel}}$, and compensation $S_{\mathrm{Comp}}$) diverged from the clinical trajectory in 
patient-specific ways, exposing which aspects of movement actually 
drove each patient's improvement, information the overall score and 
the ordinal ARAT necessarily compress away. The two findings only 
acquire their meaning together: the overall agreement validates the 
metric, and that validation is what licenses the domain divergence to 
be read as genuine recovery profile rather than measurement 
disagreement. A per-analysis-group $S_{\mathrm{overall}}$ MAD below the 
MCID was taken as evidence of longitudinal construct validity for 
that analysis group.

We illustrate this pattern in detail on the Grasp analysis group and summarise 
the remaining five analysis groups in 
Supplementary Table \ref{tab:longitudinal_mad}. For Grasp, $S_{\mathrm{overall}}$ 
tracked the clinical ARAT trajectory closely 
for both patients (MAD\,$S_{\mathrm{overall}}$ = 5.6\,pp for P10, 
5.9\,pp for P20, both well within the 15\,pp MCID), supporting longitudinal construct validity. 
$S_{\mathrm{overall}}$ itself rose by $\Delta\,S_{\mathrm{overall}} = 
+21.6$\,pp for P10 over the four sub-ceiling sessions and by 
$+23.8$\,pp for P20, against clinical ARAT changes of $+33.3$\,pp for 
both patients on the rescaled 0--100 axis. Beneath this agreement, the 
two patients showed mechanistically opposite domain profiles. P10's 
improvement was driven by gains in joint range of motion 
($\Delta\,S_{\mathrm{ROM}} = +36.1$\,pp; MAD = 5.4\,pp) and reduced 
compensatory movement ($\Delta\,S_{\mathrm{Comp}} = +32.1$\,pp; 
MAD = 6.8\,pp, mostly through reduced shoulder abduction, 
$\Delta \geq 18\,°$ from first to last session), while velocity 
remained essentially flat ($\Delta\,S_{\mathrm{Vel}} = -3.3$\,pp; 
MAD = 28.4\,pp), a range-of-motion-driven recovery without a speed 
component. P20 showed the opposite profile on the same task: velocity 
rose strongly ($\Delta\,S_{\mathrm{Vel}} = +72.3$\,pp; MAD = 16.2\,pp) 
while compensation diverged ($\Delta\,S_{\mathrm{Comp}} = -36.0$\,pp; 
MAD = 31.3\,pp), a recovery driven by faster, more vigorous execution 
with compensation worsening over the window.

The same pattern of overall concordance with patient-specific domain 
divergence held across the remaining five analysis groups 
(Supplementary Table \ref{tab:longitudinal_mad}). Evaluated per analysis group, 
the $S_{\mathrm{overall}}$ MAD fell below the MCID in all six analysis groups 
for P20 (range 5.9--13.6\,pp) and in three of six for P10 (Grasp, Hand 
to Mouth, Pinch; range 5.6--8.5\,pp). The three P10 exceedances 
(Grip 18.2\,pp; Hand Behind Head 17.4\,pp; Hand on Top of Head 
18.1\,pp) fell precisely in the analysis groups where the general-purpose 
metric set is known to be undersized: Grip, where finger kinematics are absent, and the two overhead tasks, where 
the set is optimised for reach-and-transport rather than overhead 
positioning, paralleling the metric-task mismatch flagged for Water 
Pouring at Tier~1. The exceedances thus reflect the same task-metric mismatch 
identified cross-sectionally rather than a breakdown of longitudinal 
validity, and do not affect the Grasp analysis group result on which the 
recovery contrast rests.

These case studies are consistent with H3a and H3b: the per-analysis-group 
validity criterion ($S_{\mathrm{overall}}$ MAD below the MCID, H3a) was met in both 
patients in the analysis groups where the metric set matched the task (all 
six for P20; Grasp, Hand to Mouth, and Pinch for P10). Within this 
validated frame, the patient-specific domain divergence (H3b), most cleanly 
illustrated by the opposite ROM-and-Compensation versus Velocity 
recovery profiles on Grasp, exposes recovery profiles that 
$S_{\mathrm{overall}}$ and the ordinal ARAT necessarily compress away: 
$S_{\mathrm{overall}}$ confirms what the clinical scale measures where the metric 
set matches the task, and the domain compounds explain it where 
patient-specific profiles are present. As exploratory evidence from 
two patients, these results motivate cohort-level investigation of the 
coupled finding.

\subsubsection*{Post-Ceiling: Continued Kinematic Recovery After the Clinical Scale Saturates (H3c)}
After P20 first reached the maximum mean clinical ARAT score of 3 in the Grasp analysis group at 
session~3, the clinical score remained at ceiling for all subsequent 
sessions, providing no further information about recovery. Kinematic 
metrics, however, continued to change throughout the remaining six 
sessions (see Figure~\ref{fig:Longitudinal_P20}). On the normalised 
compound scale, $S_{\mathrm{overall}}$ rose by a further 
$\Delta\,S_{\mathrm{overall}} = +12.3$\,pp across the post-ceiling 
window. Velocity metrics showed the 
strongest post-ceiling change: $S_{\mathrm{Vel}}$ 
increased by $\Delta\,S_{\mathrm{Vel}} = +24.4$\,pp 
during the post-ceiling window, exceeding the 
15\,pp MCID. 
ROM metrics showed smaller 
positive changes (shoulder flexion ROM: $+$27.8\,\% of cohort 
range, raw metric), and trunk displacement decreased modestly 
($-$8.5\,\% of cohort range, raw metric). P20 surpassed the normal motor-performance 
threshold of the velocity compound only several sessions after reaching the clinical score ceiling, indicating 
that kinematic normalisation toward the per-metric normal motor-performance was 
achieved only towards the end of monitoring, a distinction entirely 
invisible to the ordinal clinical scale.
In this single post-ceiling case, kinematic measurement therefore 
detected clinically relevant change after the ordinal scale had 
saturated, consistent with H3c.

\newpage

\section*{Discussion}
Effective and efficient neurorehabilitation requires therapy to be 
steered by what is actually measured, and precision neurorehabilitation 
has been proposed as the framework to achieve this \cite{adans-dester_enabling_2020}. Its central 
bottleneck, however, is measurement \cite{kwakkel_standardized_2019}. This study evaluated whether 
AI-based markerless motion capture, used inside routine ARAT 
assessments in a neurorehabilitation clinic, accurately and robustly 
reconstructs upper limb movement and yields kinematic metrics that are 
valid and clinically informative beyond the ordinal score, and could 
therefore help close that measurement loop at scale.

We addressed this 
through four hypotheses, spanning reconstruction accuracy (H1), 
cross-sectional construct validity (H2), and longitudinal validity, 
specificity, and beyond-ceiling responsiveness in two exploratory case 
studies (H3a--H3c). All were supported within their pre-specified 
thresholds: H1 and H2 held at the cohort level, and the longitudinal 
criterion (H3a) held in all six analysis groups for P20 and three of 
six for P10. We discuss each in turn, then the limitations and future 
directions the work opens.

Reconstruction accuracy is the precondition for every later claim, and 
it held where it is most likely to fail: despite an ad-hoc three-camera 
setup in routine sessions, reprojection error stayed within the 
acceptable clinical-use range and did not degrade with impairment, 
making the underlying signal trustworthy enough to build metrics on.

Validating kinematic measures against a clinical score is complicated 
by the obstacle Schwarz et al.\ \cite{schwarz_systematic_2019} 
identify: kinematic metrics provide complementary rather than redundant 
information, so classical convergent correlation with the ordinal scale 
cannot be the test. A known-groups approach~\cite{de_vet_measurement_2011} 
sidesteps this by asking not whether the metrics agree with the ARAT 
but whether they agree \emph{in the way a valid measure should}. They 
did: discrimination was strong where the clinical boundary is robust 
(Tier~1, completed vs.\ uncompleted) and attenuated where it is 
subjective (Tier~2, quality of completion). This differential pattern, 
rather than uniform agreement, is the signature of a continuous measure 
tracking the underlying construct and 
generalises to any setting where a continuous measure must be validated 
against a noisy ordinal reference.

The longitudinal case studies most directly motivate clinical 
deployment beyond the ordinal score. On the Grasp analysis group, the 
overall compound tracked the clinical ARAT closely in both patients, 
while their domain decompositions revealed mechanistically opposite 
recovery profiles: both gained the same ARAT points, yet one recovered 
through range of motion and reduced compensation and the other through 
faster execution. Overall agreement and domain decomposition are interpretable only 
together: the overall agreement legitimises the compound as tracking 
the same signal as the ARAT, which is what lets the domain 
decomposition be read as recovery profile rather than measurement 
noise. The same clinical improvement was thus driven by different 
movement profiles, directly addressing the third ARAT limitation.
Crucially, kinematic measurement remained responsive beyond the 
clinical ceiling: in P20, the compound captured continued, 
velocity-driven recovery across the post-ceiling window that the 
saturated ARAT could not register, and resolved its profile, not merely 
that overall improvement continued but also in which domain.

Several considerations bound these results, each with a clear path 
forward now that the measurement infrastructure is demonstrated to be 
viable. The longitudinal evidence rests on two case studies, motivating 
but not establishing the coupled finding at the cohort level; a 
prospective cohort study with structured assessment schedules is the 
natural next step. The MCID adopted here is an approximation translated 
from consensus recommendations \cite{kwakkel_standardized_2019} rather 
than estimated for this measure and cohort; a larger cohort would allow 
movement-specific cut-offs and measure-specific reliability bounds to be 
established directly, for which criterion-anchored estimation against 
patient-perceived change provides a template \cite{pohl_construct_2025}. 
ARAT scores were also assigned by treating therapists without parallel 
inter-rater reliability data, leaving the contribution of clinical 
subjectivity to the Tier~2 attenuation unquantified. A final, 
cross-cutting limitation is that the added insight depends on each 
metric's domain assignment holding for the task at hand: because every 
feature is assigned to a single domain across the full ARAT battery, the 
same feature can mean different things across tasks, and this surfaces 
wherever the assignment breaks down. Cross-sectionally, Water Pouring 
showed the lowest Tier~1 discrimination (combined-feature AUC 0.75), as 
shoulder abduction, assigned as compensation for reach-and-transport, may 
instead be a legitimate pouring strategy; longitudinally, the same 
mismatch excluded Water Pouring and drove P10's three MAD exceedances 
(Grip, lacking finger kinematics; the two overhead tasks, assigned for 
reach-and-transport rather than overhead positioning). Where a metric's 
domain assignment is loosely coupled to a task's actual recovery, the 
compound tracks the clinical trajectory less faithfully; the principled 
response is to surface such cases rather than report misleading 
agreement. This is not an accuracy limitation but a representational one: 
the present hand-crafted, domain-assigned metrics are a deliberately 
general first representation rather than the last word.

Beyond extending the hand-crafted metric set, more fundamental 
measurement-side directions follow naturally and are made viable by 
the routine acquisition demonstrated here. Scoring the whole movement by 
its deviation from an able-bodied reference, rather than reducing it 
to pre-chosen scalars, sidesteps the per-task assignment problem, 
since the reference itself encodes what healthy execution of each task 
looks like; this is established practice in gait analysis, where 
indices such as the Gait Deviation Index quantify deviation from an 
able-bodied reference set \cite{schwartz_gait_2008}, and is beginning 
to be explored for the upper limb \cite{zaidi_upper_2023}. Further 
still, because the underlying reconstruction yields full biomechanical 
trajectories rather than isolated metrics, the data are directly 
consumable by emerging movement foundation models that learn the 
shared structure of human movement and answer clinical questions, 
including outcome scoring, without task-specific feature engineering 
\cite{yang_biomechgpt_2025}. Realising these data-driven directions 
requires able-bodied and patient data at population scale, and would 
also replace the present Score~3 normal motor-performance proxy with measured 
able-bodied references; the routine, low-burden acquisition 
demonstrated here is one route to generating that volume of data. In 
parallel, the underlying tracking can be enriched along a hardware 
axis: returning to a five-camera configuration, as used in laboratory 
validation, would support reliable finger kinematics and reduce 
occlusion-driven artefacts \cite{maggioni_optimisation_2025, firouzabadi_biomechanical_2024}, 
while at the opposite extreme single-camera reconstruction 
\cite{donahue_monocular_2026, yang_sam_2026, cotton_monocular_2026} 
could lower the deployment barrier further toward ubiquitous, in-room 
measurement.
\section*{Conclusion}
The present results represent a first exploration of a new 
measurement territory in which full kinematic reconstruction during 
routine clinical assessment becomes possible through passive 
camera instrumentation. Passive 
continuous kinematic measurement during routine assessment opens a 
new mode of motor evaluation: observer-independent, 
multi-dimensional, sensitive past the clinical ceiling, capable of 
revealing recovery profiles that ordinal scoring cannot, and able 
to generate the volume and continuity of structured data needed to 
bring modern data-driven methods into routine practice and close the 
loop for precision upper-limb motor neurorehabilitation.

\bibliographystyle{IEEEtran}
\bibliography{references}

\newpage

\section*{Acknowledgements}

We thank all participants of the study. This study was enabled 
through generous funding from the P \& K Foundation.

\section*{Author contributions statement}

T.U. conceived and designed the study, developed the analysis 
pipeline, performed the data analysis, interpreted the results, and 
drafted the manuscript. C.E.A. supervised the project, contributed to 
the conception and design of the study, and revised the manuscript. 
R.J.C. provided technical guidance on the markerless motion capture 
and biomechanical reconstruction methodology and revised the 
manuscript. A.R.L. provided clinical oversight and access to the 
clinical site at which the data were collected, and revised the 
manuscript. O.L. and R.G. contributed to the scientific framing and 
methodological direction of the work and revised the manuscript. All 
authors reviewed and approved the submitted version.

\section*{Additional information}

\textbf{Competing interests} The authors declare no competing interests.

\textbf{Ethics approval} This study was approved by the cantonal ethics 
committee (EKNZ, 2024-00196). All participants provided informed 
consent for research use of their data.

\textbf{Data availability} The datasets generated and analysed in this 
study contain potentially identifiable clinical and video-derived data 
and are therefore not publicly available owing to data-protection and 
participant-privacy constraints. De-identified data that can be shared 
in compliance with the applicable data-protection regulations and 
ethical approvals are available from the corresponding author on 
reasonable request.

\newpage
\section*{Supplementary Material}

\begin{figure}[ht]
    \centering
    \includegraphics[width=0.4\textwidth]{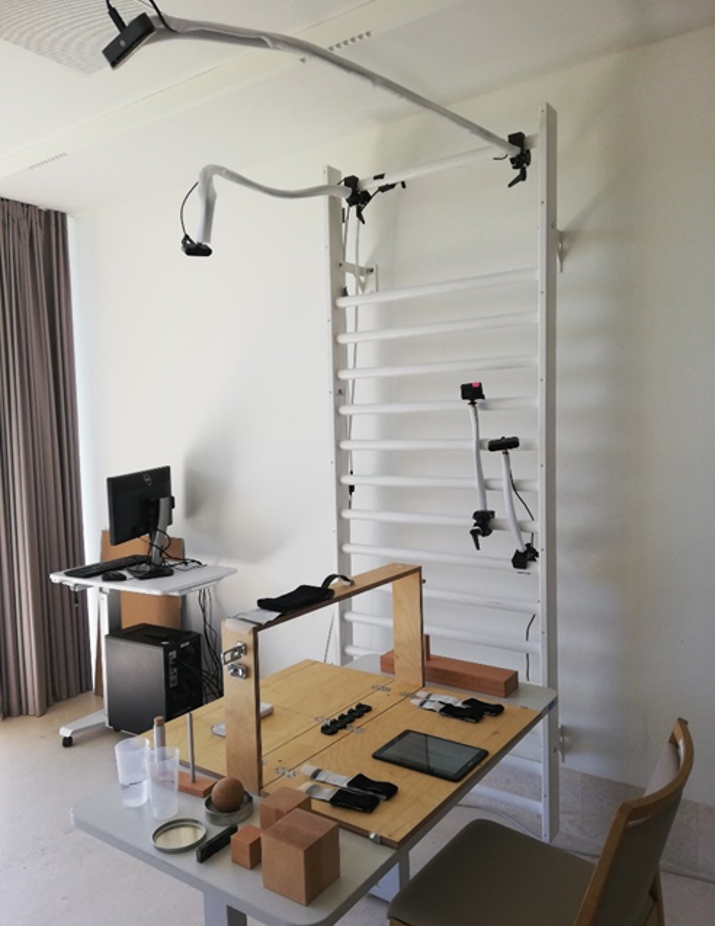}
    \caption{Ad-hoc camera setup around the ARAT assessment table.}
    \label{fig:ARAT_camera_setup}
\end{figure}

\begin{figure}[hb]
    \centering
    \includegraphics[width=0.4\textwidth]{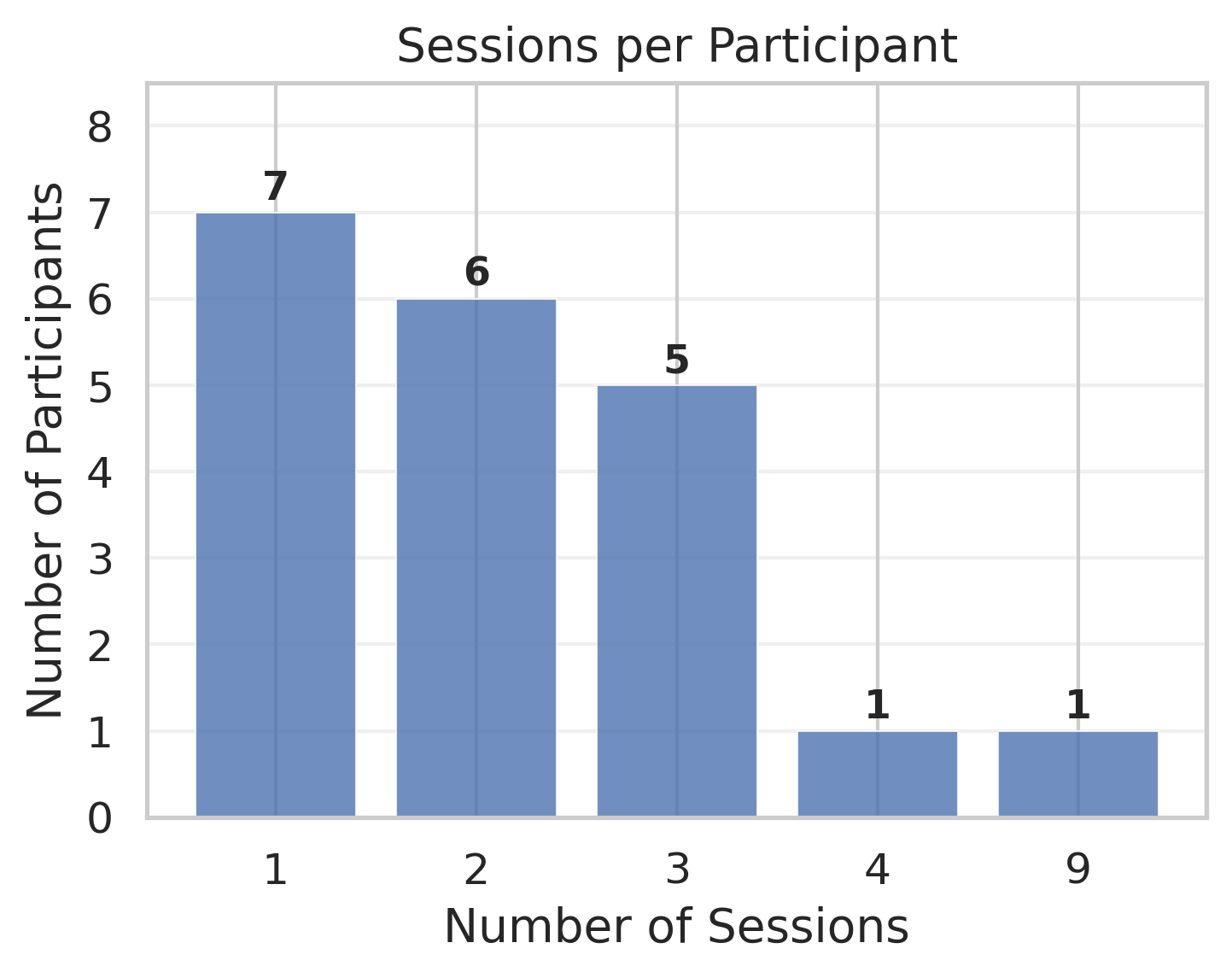}
    \caption{Sessions per participant}
    \label{fig:sessions-per-participant}
\end{figure}

\begin{figure}[b]
    \centering

    \begin{subfigure}{\linewidth}
        \centering
        \includegraphics[width=\linewidth]{results/boxplots_by_score_grasp.png}
        \caption{Grasp}
        \label{fig:grasp_boxplots_by_score_appendix}
    \end{subfigure}

    \begin{subfigure}{\linewidth}
        \centering
        \includegraphics[width=\linewidth]{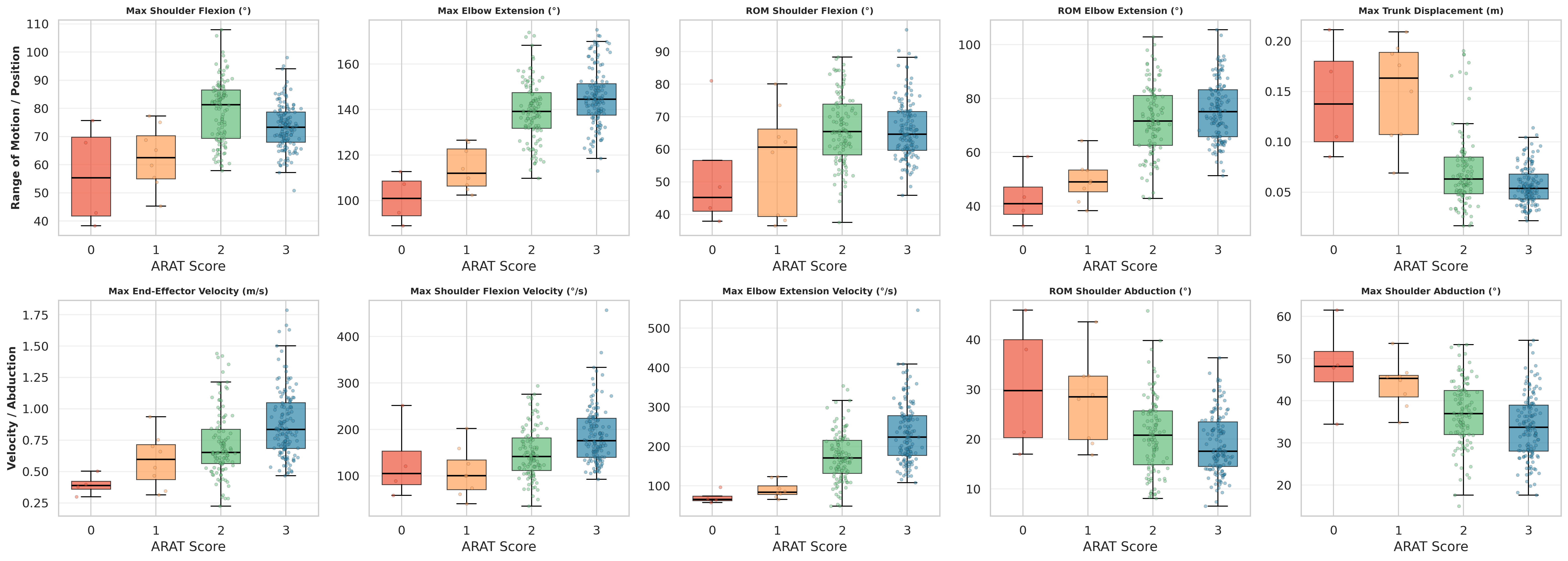}
        \caption{Grip}
        \label{fig:grip_boxplots_by_score}
    \end{subfigure}

    \begin{subfigure}{\linewidth}
        \centering
        \includegraphics[width=\linewidth]{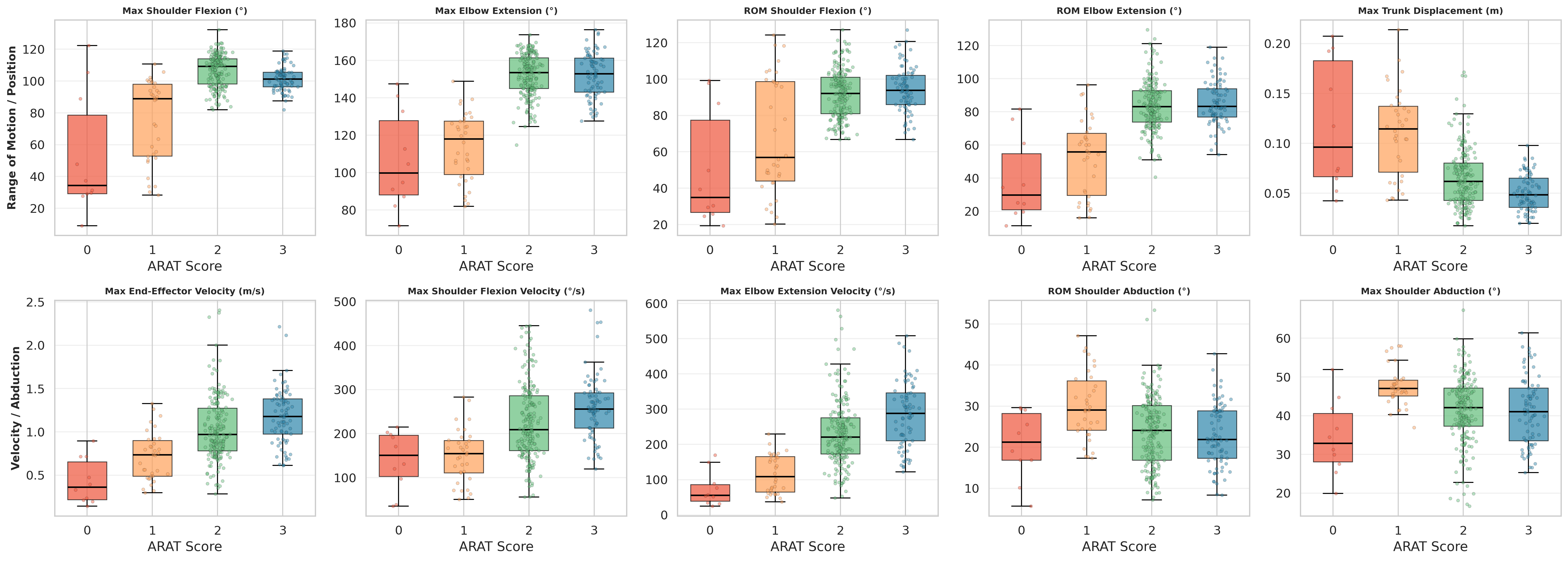}
        \caption{Pinch}
        \label{fig:pinch_boxplots_by_score}
    \end{subfigure}

    \caption{Distribution of kinematic metrics across clinical ARAT scores (0--3) for pick-and-place tasks per analysis group. Individual trials shown as jittered points. Outliers removed (IQR $\times$ 3.0).}
    \label{fig:population_boxplots_by_score}
\end{figure}

\begin{table*}
\centering
\footnotesize
\setlength{\tabcolsep}{3pt}
\caption{Population descriptive statistics across all seven task groups. Top: pick-and-place tasks (Grasp, Grip, Pinch); bottom: gross-motor and pouring tasks (Water Pouring, Hand Behind Head, Hand on Top of Head, Hand to Mouth). Raw kinematic metric rows show the median across trials in raw units (degrees, m, m/s); compound score rows show the median with the interquartile range $[Q_{25}, Q_{75}]$ on the Q75-anchored $0$--$100\,\%$ scale, where $100\,\%$ corresponds to the per-metric 75th percentile of Score~3 cohort sessions and values are not clipped at the upper bound (i.e., compound scores may exceed $100\,\%$ when performance surpasses the normal-performance threshold). Inverted compensation metrics (trunk displacement and shoulder abduction) use Q25 as the $100\,\%$ anchor so that higher normalised values always indicate better function.}
\label{tab:descriptive_stats_merged}
\begin{center}
\textit{Pick-and-place tasks} \\[2pt]
\resizebox{\linewidth}{!}{%
\begin{tabular}{ l c c c c c c c c c c c c }
\hline\hline
 & \multicolumn{4}{c}{Grasp} & \multicolumn{4}{c}{Grip} & \multicolumn{4}{c}{Pinch} \\
\cmidrule(lr){2-5} \cmidrule(lr){6-9} \cmidrule(lr){10-13}
Metric & S0 & S1 & S2 & S3 & S0 & S1 & S2 & S3 & S0 & S1 & S2 & S3 \\
 & (8) & (34) & (140) & (108) & (4) & (8) & (110) & (139) & (10) & (34) & (194) & (77) \\
\hline
$\max(\theta_{SF})$ [$^\circ$] & 53.9 & 74.8 & 104.8 & 102.2 & 55.4 & 62.5 & 81.3 & 73.3 & 34.4 & 88.9 & 109.3 & 101.3 \\
$\max(\theta_{EE})$ [$^\circ$] & 97.4 & 115.7 & 152.9 & 160.4 & 100.9 & 111.9 & 139.2 & 144.6 & 99.7 & 117.9 & 153.4 & 152.8 \\
$\max(\theta_{SA})$ [$^\circ$] & 47.6 & 47.8 & 40.3 & 36.6 & 48.1 & 45.3 & 36.9 & 33.7 & 32.9 & 47.0 & 42.1 & 41.0 \\
$\mathrm{ROM}(\theta_{SF})$ [$^\circ$] & 56.0 & 62.4 & 93.3 & 91.3 & 45.2 & 60.7 & 65.4 & 64.6 & 34.9 & 56.9 & 92.1 & 93.8 \\
$\mathrm{ROM}(\theta_{EE})$ [$^\circ$] & 33.2 & 45.9 & 80.1 & 86.8 & 40.9 & 49.0 & 71.6 & 75.1 & 29.8 & 55.8 & 83.1 & 83.3 \\
$\mathrm{ROM}(\theta_{SA})$ [$^\circ$] & 33.5 & 30.5 & 22.9 & 21.1 & 29.7 & 28.5 & 20.8 & 17.6 & 21.3 & 29.0 & 24.1 & 21.9 \\
$\max(d_{Tr})$ [m] & 0.131 & 0.062 & 0.063 & 0.055 & 0.137 & 0.163 & 0.063 & 0.054 & 0.096 & 0.114 & 0.062 & 0.048 \\
$\max(v_{\mathrm{hand}})$ [m/s] & 0.569 & 0.635 & 1.05 & 1.29 & 0.388 & 0.596 & 0.652 & 0.836 & 0.362 & 0.733 & 0.969 & 1.18 \\
$\max(\omega_{SF})$ [$^\circ$/s] & 121.9 & 130.5 & 236.2 & 280.2 & 105.0 & 100.2 & 141.7 & 175.9 & 150.8 & 154.5 & 209.1 & 255.9 \\
$\max(\omega_{EE})$ [$^\circ$/s] & 73.9 & 91.8 & 222.0 & 324.8 & 65.8 & 84.0 & 170.8 & 223.6 & 55.4 & 108.8 & 220.7 & 288.2 \\
\hline
$S_{\mathrm{ROM}}$ [\%] & 34 & 50 & 89 & 92 & 32 & 45 & 86 & 85 & 24 & 59 & 92 & 93 \\
\quad\textit{IQR} & [19, 48] & [27, 67] & [82, 97] & [87, 100] & [17, 47] & [39, 52] & [71, 100] & [75, 97] & [18, 69] & [30, 80] & [85, 100] & [85, 98] \\
$S_{\mathrm{Vel}}$ [\%] & 24 & 32 & 62 & 81 & 25 & 38 & 55 & 74 & 26 & 40 & 66 & 84 \\
\quad\textit{IQR} & [10, 43] & [13, 38] & [50, 84] & [68, 97] & [19, 33] & [21, 46] & [43, 70] & [60, 102] & [14, 39] & [25, 56] & [49, 87] & [63, 100] \\
$S_{\mathrm{Comp}}$ [\%] & 63 & 64 & 81 & 87 & 52 & 52 & 80 & 89 & 84 & 63 & 81 & 85 \\
\quad\textit{IQR} & [39, 71] & [53, 81] & [70, 91] & [78, 97] & [32, 63] & [40, 61] & [68, 92] & [79, 96] & [66, 96] & [53, 70] & [71, 90] & [74, 95] \\
$S_{\mathrm{overall}}$ [\%] & 33 & 46 & 78 & 85 & 34 & 44 & 75 & 83 & 46 & 55 & 79 & 88 \\
\quad\textit{IQR} & [32, 43] & [34, 56] & [72, 84] & [80, 94] & [30, 38] & [36, 48] & [67, 82] & [74, 94] & [40, 57] & [39, 63] & [74, 88] & [81, 95] \\
\hline\hline
\end{tabular}%
}

\vspace{3ex}
\textit{Gross-motor and pouring tasks} \\[2pt]
\resizebox{\linewidth}{!}{%
\begin{tabular}{ l c c c c c c c c c c c c c c c c }
\hline\hline
 & \multicolumn{4}{c}{Water} & \multicolumn{4}{c}{HBH} & \multicolumn{4}{c}{HOH} & \multicolumn{4}{c}{HTM} \\
\cmidrule(lr){2-5} \cmidrule(lr){6-9} \cmidrule(lr){10-13} \cmidrule(lr){14-17}
Metric & S0 & S1 & S2 & S3 & S0 & S1 & S2 & S3 & S0 & S1 & S2 & S3 & S0 & S1 & S2 & S3 \\
 & (1) & (4) & (44) & (26) & (0) & (12) & (43) & (15) & (0) & (8) & (35) & (36) & (0) & (3) & (20) & (61) \\
\hline
$\max(\theta_{SF})$ [$^\circ$] & -- & 45.3 & 64.1 & 67.7 & -- & 78.6 & 125.6 & 126.4 & -- & 99.9 & 118.3 & 129.5 & -- & 75.9 & 51.6 & 50.9 \\
$\max(\theta_{EE})$ [$^\circ$] & -- & 105.1 & 93.8 & 97.4 & -- & 135.8 & 132.0 & 117.6 & -- & 148.6 & 134.3 & 129.7 & -- & 131.5 & 143.1 & 137.3 \\
$\max(\theta_{SA})$ [$^\circ$] & -- & 32.2 & 34.0 & 30.7 & -- & 48.9 & 46.9 & 46.3 & -- & 55.8 & 46.2 & 45.7 & -- & 41.7 & 21.5 & 18.2 \\
$\mathrm{ROM}(\theta_{SF})$ [$^\circ$] & -- & 44.6 & 52.7 & 58.3 & -- & 72.0 & 120.2 & 124.1 & -- & 95.1 & 114.3 & 120.5 & -- & 78.9 & 36.2 & 33.9 \\
$\mathrm{ROM}(\theta_{EE})$ [$^\circ$] & -- & 30.9 & 18.8 & 16.1 & -- & 101.4 & 92.8 & 83.8 & -- & 91.0 & 84.0 & 75.0 & -- & 95.7 & 102.9 & 94.3 \\
$\mathrm{ROM}(\theta_{SA})$ [$^\circ$] & -- & 22.6 & 15.0 & 11.0 & -- & 41.0 & 34.2 & 30.9 & -- & 48.2 & 33.3 & 30.5 & -- & 32.4 & 19.1 & 22.2 \\
$\max(d_{Tr})$ [m] & -- & 0.060 & 0.044 & 0.049 & -- & 0.050 & 0.030 & 0.028 & -- & 0.044 & 0.028 & 0.026 & -- & 0.046 & 0.024 & 0.017 \\
$\max(v_{\mathrm{hand}})$ [m/s] & -- & 0.609 & 0.463 & 0.591 & -- & 1.06 & 1.93 & 2.45 & -- & 1.20 & 1.81 & 2.45 & -- & 1.06 & 1.15 & 1.69 \\
$\max(\omega_{SF})$ [$^\circ$/s] & -- & 117.7 & 99.9 & 126.0 & -- & 212.4 & 365.6 & 459.7 & -- & 211.6 & 334.0 & 398.4 & -- & 135.8 & 94.7 & 142.6 \\
$\max(\omega_{EE})$ [$^\circ$/s] & -- & 134.2 & 47.4 & 58.7 & -- & 210.8 & 285.5 & 286.5 & -- & 170.2 & 217.1 & 305.0 & -- & 197.2 & 229.7 & 314.5 \\
\hline
$S_{\mathrm{ROM}}$ [\%] & -- & 86 & 74 & 80 & -- & 89 & 103 & 84 & -- & 73 & 85 & 77 & -- & 135 & 85 & 76 \\
\quad\textit{IQR} & -- & [51, 119] & [59, 104] & [63, 91] & -- & [42, 105] & [87, 118] & [77, 95] & -- & [63, 102] & [69, 106] & [71, 93] & -- & [131, 167] & [74, 99] & [65, 89] \\
$S_{\mathrm{Vel}}$ [\%] & -- & 125 & 53 & 85 & -- & 37 & 73 & 81 & -- & 36 & 58 & 81 & -- & 41 & 46 & 78 \\
\quad\textit{IQR} & -- & [92, 135] & [42, 73] & [70, 98] & -- & [27, 49] & [51, 87] & [77, 95] & -- & [28, 46] & [31, 76] & [66, 93] & -- & [39, 57] & [26, 77] & [62, 102] \\
$S_{\mathrm{Comp}}$ [\%] & -- & 77 & 87 & 94 & -- & 76 & 97 & 97 & -- & 60 & 83 & 87 & -- & 53 & 86 & 90 \\
\quad\textit{IQR} & -- & [63, 85] & [71, 97] & [80, 100] & -- & [63, 97] & [81, 105] & [91, 99] & -- & [49, 64] & [71, 95] & [79, 98] & -- & [27, 58] & [74, 95] & [85, 98] \\
$S_{\mathrm{overall}}$ [\%] & -- & 92 & 74 & 84 & -- & 70 & 88 & 87 & -- & 52 & 73 & 84 & -- & 77 & 74 & 82 \\
\quad\textit{IQR} & -- & [73, 105] & [64, 84] & [76, 94] & -- & [59, 74] & [83, 93] & [83, 95] & -- & [49, 71] & [62, 84] & [74, 89] & -- & [76, 84] & [61, 81] & [75, 89] \\
\hline\hline
\end{tabular}%
}
\end{center}
\end{table*}
\begin{table}[h!]
\caption{Mean absolute deviation (MAD, in percentage points on the shared Q75-anchored 0--100\,\% scale) between change vectors of the clinical ARAT score and the kinematic compound scores across the sub-ceiling window of each patient, for all six Analysis Groups. On this scale, $100\,\%$ corresponds to the per-metric 75th percentile of Score~3 cohort sessions and values are not clipped at the upper bound; the Compensation domain is inverted during normalisation (Q25 used as the $100\,\%$ anchor instead of Q75) so that higher values consistently indicate less compensatory movement. Smaller MAD indicates closer agreement between the kinematic compound and the clinical trajectory; larger MAD indicates divergence in either magnitude or direction. $S_{\mathrm{overall}}$ = overall compound score; $S_{\mathrm{ROM}}$, $S_{\mathrm{Vel}}$, $S_{\mathrm{Comp}}$ = domain compound scores. Reference threshold: $15\,\%$ = minimal clinically important difference (MCID) \cite{kwakkel_standardized_2019}, expressed in pp on the same Q75-anchored scale. Sessions $n$ is the number of sub-ceiling sessions per task group.}
\label{tab:longitudinal_mad}
\centering
\small
\renewcommand{\arraystretch}{1.1}
\begin{tabular}{ l c c r r r r }
\hline\hline
Task & Sessions & $n$ & $S_{\mathrm{overall}}$ & $S_{\mathrm{ROM}}$ & $S_{\mathrm{Vel}}$ & $S_{\mathrm{Comp}}$ \\
     &          &     & (\% pts) & (\% pts) & (\% pts) & (\% pts) \\
\hline
\multicolumn{7}{l}{\textit{P10 --- Ischaemic stroke (ICD I63.1), right side (affected)}} \\
Grasp & 1-4 & 4 & 5.6 & 5.4 & 28.4 & 6.8 \\
Grip & 1-4 & 4 & 18.2 & 10.2 & 27.0 & 26.6 \\
Pinch & 1-4 & 4 & 8.5 & 6.2 & 12.5 & 11.1 \\
Hand Behind Head & 1-4 & 4 & 17.4 & 44.0 & 24.3 & 38.6 \\
Hand on Top of Head & 1-3 & 3 & 18.1 & 24.5 & 20.6 & 9.1 \\
Hand to Mouth & 1-3 & 3 & 7.1 & 8.1 & 10.9 & 13.3 \\
\hline
\multicolumn{7}{l}{\textit{P20 --- Stroke (right hemisphere), left side (affected)}} \\
Grasp & 1-3 & 3 & 5.9 & 3.8 & 16.2 & 31.3 \\
Grip & 1-3 & 3 & 6.3 & 3.9 & 2.1 & 14.4 \\
Pinch & 1-3 & 3 & 12.1 & 13.4 & 4.0 & 21.8 \\
Hand Behind Head & 1-3 & 3 & 13.6 & 12.3 & 39.2 & 1.4 \\
Hand on Top of Head & 1-3 & 3 & 8.0 & 17.7 & 17.9 & 12.8 \\
Hand to Mouth & 1-3 & 3 & 7.2 & 22.5 & 10.6 & 15.3 \\
\hline\hline
\end{tabular}
\end{table}

\begin{table}[h]
    \centering
    \begin{tabular}{ll}
        \toprule
        Characteristic & Value \\
        \midrule
        Age (years), mean $\pm$ SD & $71.2 \pm 9.9$ \\
        Age range (years) & 55--90 \\
        Sex: male & 15 (75\%) \\
        Height (cm), mean $\pm$ SD & $174.1 \pm 6.5$ \\
        Weight (kg), mean $\pm$ SD & $74.9 \pm 12.5$ \\
        \midrule
        Diagnosis: stroke & 10 (50\%) \\
        Diagnosis: Parkinson's disease & 6 (30\%) \\
        Diagnosis: other & 4 (20\%) \\
        \bottomrule
    \end{tabular}
    \caption{Participant characteristics ($N$=20).}
    \label{tab:demographics}
\end{table}

\end{document}